\documentclass{ieeeaccess}
\usepackage{cite}
\usepackage{amsmath,amssymb,amsfonts}
\usepackage{textcomp}

\usepackage{graphicx}%
\usepackage{multirow}%
\usepackage{amsthm}%
\usepackage{mathrsfs}%
\usepackage{xcolor}%
\usepackage{textcomp}%
\usepackage{manyfoot}%
\usepackage{booktabs}%
\usepackage{algorithm}%
\usepackage{algorithmicx}%
\usepackage{algpseudocode}%
\usepackage{listings}%

\usepackage{lineno}
\usepackage{tabularx}
\usepackage{adjustbox, color, colortbl}
\usepackage{mathtools}
\usepackage{subfig}
\usepackage{diagbox}
\definecolor{Gray}{gray}{0.9}
\usepackage{tabulary}
\usepackage{balance}
\usepackage{soul}
\usepackage{hyperref}
\newcommand{\red}[1]{{\textcolor{red}{#1}}}

\newcommand{\hlc}[2][yellow]{{%
    \colorlet{foo}{#1}%
    \sethlcolor{foo}\hl{#2}}%
}

\def\BibTeX{{\rm B\kern-.05em{\sc i\kern-.025em b}\kern-.08em
    T\kern-.1667em\lower.7ex\hbox{E}\kern-.125emX}}
\begin{document}
\history{Received 23 June 2024, accepted 17 July 2024, date of publication xx xxxxx 2024, date of current version 20 July 2024.}
\doi{xx.xxxx/ACCESS.xxxx.xxxxxxx}

\title{The Art of Camouflage: Few-shot Learning for
Animal Detection and Segmentation}
\author{\uppercase{Thanh-Danh Nguyen}\authorrefmark{1,2,5,*}, \uppercase{Anh-Khoa Nguyen Vu}\authorrefmark{1,2,5,*}, \uppercase{Nhat-Duy Nguyen}\authorrefmark{1,2,5,*}, \uppercase{Vinh-Tiep Nguyen}\authorrefmark{1,2,5}, \uppercase{Thanh Duc Ngo}\authorrefmark{1,2,5}, \uppercase{Thanh-Toan Do}\authorrefmark{6}, \uppercase{Minh-Triet Tran}\authorrefmark{3,4,5}, \uppercase{Tam V. Nguyen}\authorrefmark{7},
\IEEEmembership{Senior Member, IEEE}}

\address[1]{Laboratory of Multimedia Communications, University of Information Technology, Ho Chi Minh City, Vietnam}
\address[2]{Faculty of Computer Science, University of Information Technology, Ho Chi Minh City, Vietnam}
\address[3]{Faculty of Information Technology, University of Science, Ho Chi Minh City, Vietnam}
\address[4]{John von Neumann Institute, VNU-HCM, Vietnam}
\address[5]{Vietnam National University, Ho Chi Minh City, Vietnam}
\address[6]{Faculty of Information Technology, Monash University, Clayton, VIC 3800, Australia}
\address[7]{Department of Computer Science, University of Dayton, Dayton, OH 45469, United States}



\markboth
{Nguyen \headeretal: The Art of Camouflage: Few-shot Learning for
Animal Detection and Segmentation}
{Nguyen \headeretal: The Art of Camouflage: Few-shot Learning for
Animal Detection and Segmentation}

\corresp{Corresponding author: Tam V. Nguyen (e-mail: tamnguyen@udayton.edu), *equal contribution}

\begin{abstract}

Camouflaged object detection and segmentation is a new and challenging research topic in computer vision. There is a serious issue of lacking data on concealed objects such as camouflaged animals in natural scenes. In this paper, we address the problem of few-shot learning for camouflaged object detection and segmentation. To this end, we first collect a new dataset, CAMO-FS, for the benchmark. \hlc[white]{As camouflaged instances are challenging to recognize due to their similarity compared to the surroundings, we guide our models to obtain camouflaged features that highly distinguish the instances from the background. In this work, we propose FS-CDIS, a framework to efficiently detect and segment camouflaged instances via two loss functions contributing to the training process. Firstly, the instance triplet loss with the characteristic of differentiating the anchor, which is the mean of all camouflaged foreground points, and the background points are employed to work at the instance level. Secondly, to consolidate the generalization at the class level, we present instance memory storage with the scope of storing camouflaged features of the same category, allowing the model to capture further class-level information during the learning process.} The extensive experiments demonstrated that our proposed method achieves state-of-the-art performance on the newly collected dataset. Code is available at \href{https://github.com/danhntd/FS-CDIS}{\textcolor{magenta}{https://github.com/danhntd/FS-CDIS}}.

\end{abstract}

\begin{keywords}
Camouflaged animals, camouflaged instances, few-shot learning, object detection, and instance segmentation.
\end{keywords}

\titlepgskip=-21pt

\maketitle

\section{Introduction}
\label{sec:introduction}


Camouflage is a defense mechanism that animals use to conceal their appearance by blending in with their environment \cite{Sujit-ICEECS2013}. Autonomously detecting camouflaged animals is helpful in various applications, e.g., search-and-rescue missions~\cite{ltnghia-CVIU2019}; wild species discovery and preservation activities~\cite{ltnghia-CVIU2019}; and media forensics~\mbox{\cite{zhou2022immune,yu2024deep,yu2022floating,wang2023non}}(manipulated image/video detection and segmentation~\cite{ltnghia-ICCV2021}). 
\hlc[white]{By leveraging camouflage recognition at object detection or instance segmentation level autonomously, these practical applications can be done with minor efforts from humans while maintaining work performance. To be specific, utilizing a drone flying around the mountainous area to collect images and videos for a system to detect and segment the target object in danger is more effective than sending a group of humans manually scanning the zone. By this means, this process can support biological scientists in identifying and preserving endangered species effectively. Further related applications can be considered in different important fields including healthcare, agriculture, or military, where exist the concept of finding objects with camouflaged features. 
Indeed, camouflage detection and segmentation tasks can provide further applications such as assisting doctors in medical imaging understanding, supporting modern farmers in managing the vast fields of crops via visual information or even detecting hidden enemies in nature. These applications can be potential via the development of camouflaged research via image understanding at detailed levels of object detection and instance segmentation.} 

\begin{table*}[!t]
\caption{{\color{black} Statistics of camouflage datasets (without non-camo images).
}}
\label{table:camo_data_statistic}
\centering
\adjustbox{max width=\textwidth}{

\begin{tabular}{|l|c|c|c|c|c|c|c|c|c|c|c|}
\hline
\textbf{Dataset} & \textbf{Year} & \textbf{Venue} & \textbf{Type} & \multicolumn{1}{c|}{\begin{tabular}[c]{@{}c@{}}\textbf{\#Annot.} \\ \textbf{Camo. Img.}\end{tabular}} & \multicolumn{1}{c|}{\begin{tabular}[c]{@{}c@{}}\textbf{\#Meta-} \\ \textbf{Cat.}\end{tabular}} & \multicolumn{1}{c|}{\begin{tabular}[c]{@{}c@{}}\textbf{\#Obj.} \\ \textbf{Cat.}\end{tabular}} & 
\multicolumn{1}{c|}{\begin{tabular}[c]{@{}c@{}}\textbf{{\color{red} \#Ins.} or \#Obj.} \\ \textbf{per Img.}\end{tabular}} & 
\multicolumn{1}{c|}{\begin{tabular}[c]{@{}c@{}}\textbf{Bbox.} \\ \textbf{GT}\end{tabular}} & \multicolumn{1}{c|}{\begin{tabular}[c]{@{}c@{}}\textbf{Obj.} \\ \textbf{Mask GT}\end{tabular}}
& \multicolumn{1}{c|}{\begin{tabular}[c]{@{}c@{}}\textbf{Ins.} \\ \textbf{Mask GT}\end{tabular}} & \color{red}\textbf{Few-shot} \\ \hline
CamouflagedAnimals \cite{Bideau-ECCV2016} & 2016 & ECCV & Video 
& 181 & - & 6 & {\color[HTML]{FF0000} 1.238} & $\times$ & $\textbf{\checkmark}$ & $\textbf{\checkmark}$ & $\times$ \\
MoCA \cite{Lamdouar-ACCV2020} & 2020 & ACCV & Video & 7,617 & - & 67 & 1.000 & $\textbf{\checkmark}$ & $\times$ & $\times$ & $\times$ \\
CHAMELEON \cite{Skurowski-2018} & 2018 & - & Image & 76 & - & - & 1.000 & $\times$ & $\textbf{\checkmark}$ & $\times$ & $\times$ \\
CAMO \cite{ltnghia-CVIU2019} & 2019 & CVIU & Image & 1,250 & 2 & 8 & 1.000 & $\times$ & $\textbf{\checkmark}$ & $\times$ & $\times$ \\
COD \cite{fan2020camouflaged} & 2020 & CVPR & Image & 5,066 & 5 & 69 & {\color[HTML]{FF0000} 1.171} & $\textbf{\checkmark}$ & $\textbf{\checkmark}$ & $\textbf{\checkmark}$ & $\times$ \\
NC4K \cite{lv2021simultaneously} & 2021 & CVPR & Image & 4,121 & 5 & 69 & {\color[HTML]{FF0000} 1.171} & $\textbf{\checkmark}$ & $\textbf{\checkmark}$ & $\textbf{\checkmark}$ & $\times$ \\
CAMO++ \cite{le2021camouflaged} & 2022 & TIP & Image & 2,695 & 10 & 47 & {\color[HTML]{FF0000} 1.171} & $\textbf{\checkmark}$ & $\textbf{\checkmark}$ & $\textbf{\checkmark}$ & $\times$ \\ \hline
\rowcolor{Gray}
CAMO-FS (Ours) & 2024 & IEEE ACCESS & Image & 2,852 & 10 & 47 & {\color[HTML]{FF0000} 1.172} & $\textbf{\checkmark}$ & $\textbf{\checkmark}$ & $\textbf{\checkmark}$ & {\color{red} $\textbf{\checkmark}$} \\ \hline
\end{tabular}}
\end{table*}

Although image segmentation methods have been proposed for a long time, general detectors cannot deal with camouflaged animals~\cite{Kervrann-TIP1995, Boykov-IJCV2006, Li-ICASSP2011, Sulimowicz-ICIP2018} \hlc[white]{due to their specific camouflaged features}. The detectors initially developed for camouflage detection~\cite{Galun-ICCV2003, song2010new, Xue-MTA2016, Pan-MAS2011, Liu-TIP2012, Sengottuvelan-ICETET2008, Yin-PE2011, Gallego-ICIP2014}, which use handcrafted low-level features, are effective only for images with a simple and uniform background. More recently developed deep learning-based detectors~\cite{ltnghia-CVIU2019,zhai2021mutual,li2021uncertainty,fan2020camouflaged,mei2021camouflaged, Jinnan-IEEEAccess2021, Jinchao-AAAI2021, lv2021simultaneously, Skurowski-2018} for camouflaged object segmentation. 
\hlc[white]{Most previous methods are trained on large-scale datasets to perform computer vision tasks. Nevertheless, building such standard datasets for camouflaged objects is labor-intensive due to the ambiguity between the objects and their backgrounds, which leads to more time and cost in the labeling procedure.}
\hlc[white]{To address the problem, we consider the camouflaged object detection and instance segmentation under the few-shot learning which shows potential results when utilizing limited labeled samples to classify~\mbox{\cite{vinyals2016matching, finn2017model, cai2018memory, afrasiyabi2022matching, rajasegaran2020self, ma2021partner, rizve2021exploring, xing2022rethinking}}, detect~\mbox{\cite{TFA, defrcn, zhu2021semantic, yolo-reweighting, meta-rcnn, rpn-attention, meta-detr, han2022few, bulat2023fs, hu2021dense}}, or even segment~\mbox{\cite{Ganea_2021_CVPR,vu2023instance, lang2023base, cheng2022holistic, lang2024few}} new objects. 
However, to the best of our knowledge, there is a lack of camouflaged datasets supporting few-shot learning in camouflaged research.
Therefore, we introduce a new benchmark, CAMO-FS, for camouflaged few-shot object detection and instance segmentation under the few-shot settings. The new benchmark is mainly reconstructed from the CAMO++ dataset~\mbox{\cite{le2021camouflaged}} due to its diversity. The new benchmark consists of $197$ camouflaged images for training and $2,655$ camouflaged images for performance evaluation as described in} \red{Table~\ref{table:camo_data_statistic}}.

\hlc[white]{There are several approaches for few-shot learning. Beginning with few-shot classification (FSL), many works are based on meta-learning~\mbox{\cite{vinyals2016matching, finn2017model, cai2018memory, afrasiyabi2022matching, chen2023semantic}} or transfer-learning~\mbox{\cite{rajasegaran2020self, ma2021partner, rizve2021exploring, xing2022rethinking}} approaches to leverage a few labeled data to classify new objects and achieve incredible results. The former approaches compute the similarity between query and support images to pinpoint novel objects while the latter involves utilizing knowledge from the source domain to adapt a different but related target domain. Fueled by these successes, most existing works on few-shot object detection (FSOD) and few-shot segmentation (FSS) which are recently developed to tackle the problem through meta-learning~\mbox{\cite{yolo-reweighting, meta-rcnn, rpn-attention, meta-detr, han2022few, bulat2023fs, hu2021dense, sun2021fsce, max-margin, trans-int, lang2023base, cheng2022holistic, lang2024few}} and transfer-learning~\mbox{\cite{Ganea_2021_CVPR, TFA, defrcn, zhu2021semantic, li2023disentangle, vu2023instance, xu2023generating}} methods. Nonetheless, such methods focus on the general domain and thus fail to generate effective features for camouflaged objects due to the ambiguity between backgrounds and foregrounds.


To overcome the specific issue of camouflage objects, we propose a novel framework for few-shot camouflaged object detection and instance segmentation, dubbed FS-CDIS, which is based on the transfer-learning approach. The model is trained} on two stages of processing: (1) one base phase training for the model to gain concept knowledge of general domains with abundant data, and then (2) performing a novel phase that can do the specific task on the few-shot data. \hlc[white]{To be specific, in the base training stage, we train our model on generic object detection and instance segmentation dataset (e.g. COCO~\mbox{\cite{lin2014microsoft}}) and focus on improving the model in the novel fine-tuning stage. Because of the similarity between camouflaged objects and their surroundings, we aim to guide our few-shot models to obtain camouflaged features that highly distinguish the instances from the background. To achieve that target, we introduce two loss functions contributing directly to the novel fine-tuning process. Firstly, the instance triplet loss with the characteristic of differentiating the anchor, which is the mean of all camouflaged foreground points, and the background points are employed to work at the instance level. Secondly, to consolidate the generalization at the class level, we present instance memory storage with the scope of storing camouflaged features of the same category, allowing the model to capture further class-level information during the learning process.}

To summarize, our contributions in this work are two-fold:
\begin{itemize}
\item First, we build a new benchmark dataset, CAMO-FS, which is among the first datasets to support detection and instance segmentation on camouflaged instances in nature under the few-shot concept.
\item Second, we propose a framework to detect and segment camouflaged instances efficiently, named after FS-CDIS, given a small shot of training data for novel classes utilizing the idea of instance triplet loss and instance memory storage. 
\end{itemize}

The remainder of this paper is organized as follows. \textcolor{red}{Section} \ref{related_work} summarizes related work. Next, \textcolor{red}{Section} \ref{proposed_method} introduces the newly constructed CAMO-FS dataset and presents our proposed framework for few-shot camouflaged object detection and segmentation. \textcolor{red}{Section} \ref{experiments} presents the experimental results and comparison among baselines and our proposals on the newly constructed dataset. Finally, \textcolor{red}{Section} \ref{conclusion} summarizes the key points and mentions future work.

\section{Related Work}
\label{related_work}
\subsection{Camouflage Research}
Given any region (i.e. bounding boxes or polygon masks) presented for an object of interest (i.e. animals or artificial objects) in an image and then they tend to be classified as background, contents in that region can be qualified as camouflaged objects. Thus,
a camouflaged object is defined as a set of bounding boxes or camouflaged pixels in an image without any further detailed information such as the number of objects or the semantic meaning~\cite{ltnghia-CVIU2019}. \hlc[white]{Although tasks related to camouflaged animals can be performed in a wide range of applications such as security systems~\mbox{\cite{zhou2022immune,yu2024deep}}, pollution detection~\mbox{\cite{yu2022floating}}, watermark detection~\mbox{\cite{wang2023non}}}, this research field has not been well explored in the literature, especially few-shot learning which is practically suitable to the context of scare data as camouflaged animals.   

\noindent\textbf{Binary camouflage segmentation.} Prior to the advancement of deep neural networks, most of the work exploits identical regions between camouflaged regions and the background by handcrafted or low-level features, specifically based on external characteristics (e.g., color, shape, orientation, and brightness). Particularly, early camouflage detection works had attention on the foreground region even when some of its texture was similar to the background~\cite{Galun-ICCV2003, song2010new, song2010new, Xue-MTA2016}. The foreground was distinguishable from the background via simple features, such as color, intensity, shape, orientation, and edge~\cite{siricharoen2010robust,Galun-ICCV2003,kavitha2011efficient,song2010new,Xue-MTA2016}. A few methods ~\cite{Pan-MAS2011, Liu-TIP2012, Sengottuvelan-ICETET2008, Yin-PE2011, Gallego-ICIP2014} based on handcrafted low-level features have been proposed for tackling the problem of camouflage detection. However, they are effective only for images with a simple and uniform background. Thus, their performances are unsatisfactory in camouflaged object segmentation due to the substantial similarity between the foreground and the background. 

\noindent Until now, the convention of binary prefers binary ground truth camouflaged object datasets~\cite{fan2020camouflaged,ltnghia-CVIU2019,Skurowski-2018}. Existing methods for camouflaged objects~\cite{ltnghia-CVIU2019,zhai2021mutual,li2021uncertainty,fan2020camouflaged,mei2021camouflaged, Jinnan-IEEEAccess2021, Jinchao-AAAI2021, lv2021simultaneously} based on binary ground truth are considered as the binary camouflage segmentation. For example, Le \textit{et al.} \cite{ltnghia-CVIU2019} proposed an end-to-end Anabranch Network, dubbed ANet which includes two streams of classification and segmentation. The outputs of both streams are fused to improve the segmentation performance of camouflaged objects. This proposed network was also flexibly applied to any fully convolutional networks. Similarly, motivated by the way of hunting strategies of predators, Fan \textit{et al.} \cite{fan2020camouflaged} designed Search Identification Network (SINet) with two main modules to simulate this hunting behavior, namely a search module searching for targets and an identification module identifying the existence of targets then catching them.
Yan \textit{et al.} \cite{Jinnan-IEEEAccess2021} recently introduced MirrorNet, a dual-stream network comprising a mainstream and a mirror stream. This mirror stream aimed to capture instinct information by horizontally flipping camouflaged objects to break their camouflaged nature and make them more distinguishable. Zhu \textit{et al.} \cite{Jinchao-AAAI2021} presented the TINet, which interactively refines multi-level texture and segmentation features and thereby gradually enhances the segmentation of camouflaged objects. {\color{black} Lv \textit{et al.} \cite{lv2021simultaneously} simultaneously worked on ranking and localization to well-present camouflaged objects. As a result, they formed a triplet task with localizing, segmenting, and ranking the camouflaged objects. Besides, the authors also introduced the NC4K dataset for camouflaged segmentation}. {\color{black}Such methods reveal the presence of the camouflaged objects with the high level of bounding boxes and contain corresponding pixel-wise ground truth belonging to camouflage}. Further understanding of the camouflage level may help us to give comparative analyses, finding evidence for links between camouflage and other defensive strategies with aspects of habitat and life-history \cite{price2019background}.

{\color{black}\noindent\textbf{Camouflage instance segmentation.}}   {\color{black} Although several works have been proposed, there is still a difficulty in efficiently exploring the information of camouflage animals, especially at the instance level with more challenging detailed masks. Therefore,
for ease of training methods with the challenging
task of camouflaged instance segmentation, Le \textit{et al.} \cite{le2021camoufinder} introduced a framework with several state-of-the-art methods and proposed a tool with user interactive cues to tune the segmentation mask on a website. Realizing that the semantic level is not detailed enough, Le \textit{et al.} \cite{le2021camouflaged} introduced a camouflage fusion learning (CFL) to utilize the strength of different instance segmentation methods by fusing various models via learning image contexts.}

\noindent\textbf{Camouflage datasets.}
CamouflagedAnimals~\cite{Bideau-ECCV2016} and CHAMELEON~\cite{Skurowski-2018} were the first two camouflage datasets with mask annotations. The two datasets do not contain enough images to train deep learning methods. Le \textit{et al.}~\cite{ltnghia-CVIU2019} created the CAMO dataset, the first camouflage dataset with more than $1,000$ annotated images. It contains $1,250$ annotated images, which is a limited number of samples to train and evaluate deep learning methods. Then, Fan \textit{et al.}~\cite{fan2020camouflaged} collected the COD dataset, which comprises $10,000$ images (both camouflage and non-camouflage) divided into 5 meta-categories. However, they annotated only $5,066$ camouflage images. Lamdouar \textit{et al.}~\cite{Lamdouar-ACCV2020} recently developed the MoCA dataset for the camouflage object detection task; it contains only bounding box ground truths. Hence, these datasets limit their annotations at binary ground truth datasets which have a shortage of intensive annotations for multi-task camouflage problems. CAMO++ \cite{le2021camouflaged} is different from the aforementioned datasets providing a benchmark for camouflaged instance segmentation with more comprehensive annotations and diverse meta-categories of 10. The dataset comprises $5,500$ images with superiority over other datasets on instances including $32,756$ instances for both camo and non-camo objects. \hlc[white]{Different from the existing work, we address camouflage research under the few-shot learning concept to detect objects and segment camouflaged instances. Therefore, we introduce a new benchmark, dubbed CAMO-FS, to support the evaluation process of this specific task. Accordingly, our CAMO-FS comprises $2,852$ images as a result of the inheritance from CAMO++ ~\mbox{\cite{le2021camouflaged}} and our further collection. This specific dataset serves $93,1\%$ of the annotated images for the evaluation process while the rest few samples are provided for training (i.e. $197$ images for training and $2,655$ images for testing).}

\subsection{Few-shot Learning}
\textbf{Few-shot object detection (FSOD).}
When having some available samples of given classes with their corresponding bounding boxes, FSOD aims to learn from these limited data in order to help models adapt to the new classes. To date, several works \cite{rpn-attention, yolo-reweighting, TFA, meta-rcnn} have been proposed to deal with FSOD. Early works \cite{yolo-reweighting, meta-rcnn} mainly prefer to overcome the difficulties of the data scarcity of FSOD via meta-learning approaches by combining supportive information from meta-based streams with their main streams. Particularly, Bingyi \cite{yolo-reweighting} proposed a Feature Reweighting framework that leverages the free-proposal approach of a well-known one-stage framework such as YOLO~\cite{yolov2} to boost FSOD performance. The network integrated a meta-model that aims to generate reweighting vectors from support samples for highlighting the attention to features from the YOLO network. Conversely, Meta RCNN \cite{meta-rcnn} based on the two-stage proposal approach as Mask RCNN \cite{Kaiming-ICCV2017} and fed available annotations such as bounding boxes and segmented masks to train a meta-network called Predictor-head Remodeling Network for inferring attention features. Fan \textit{et al.} \cite{rpn-attention} recently proposed to take advantage of support images from a massive FSOD dataset to generate significant results combined with their proposed network called Attention-RPN, Multi-Relation Detectors. The Attention-RPN directed the trained model to look at the image for the task of object detection. Differently, Wang \textit{et al.} \cite{TFA} simply adopted Faster RCNN with two-stage finetuning to transfer massive knowledge from abundant data in the base model to fine-tune the novel one by freezing the whole network except for the fully connected layer for object classification. Through this simple straightforward mechanism, this model significantly improved few-shot performance without a complex pipeline of training the model.
Further, such works \cite{max-margin, hu2021dense, xiao2020few, han2021query, han2022meta} presented advanced methods by applying class max-margin, multiple scale proposals, or feature alignment in FSOD. Other ones were based on transformed inputs \cite{trans-int, wu2020multi}, transformer approaches \cite{meta-detr, han2022few}, contrastive method \cite{sun2021fsce}, or kernels design \cite{zhang2022kernelized}. Other methods \cite{TFA, defrcn, zhang2021hallucination, zhu2021semantic} relied only on query images to deal with FSOD via extra text data \cite{zhu2021semantic}, unlabeled image \cite{khandelwal2021unit}, generated samples \cite{zhang2021hallucination}, gradient scaling \cite{defrcn}.

\begin{figure}[!t]
\centering
    \includegraphics[width=0.49\textwidth]{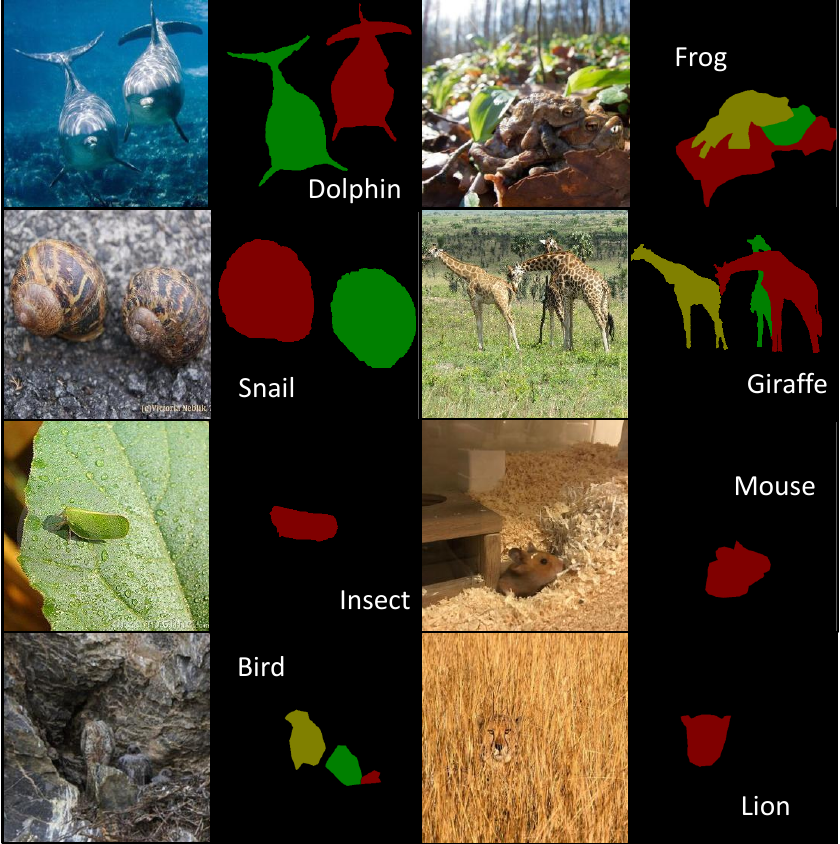}
\caption{Exemplary images with instance-level mask annotations from our proposed CAMO-FS dataset.}
\label{fig:data_sample}
\end{figure}

\noindent\textbf{Few-shot object segmentation (FSS).} 
Recently, the field of few-shot segmentation gained attention from the community. As mentioned above, the first work Meta RCNN originated from Mask RCNN, therefore, Meta RCNN simultaneously performed detection and segmentation. Liu \textit{et al.} \cite{Liu_2020_CVPR} utilized a cross-reference network for generic image segmentation. The authors proposed a cross-reference mechanism and a mask refinement module to specifically support the task of segmentation. Before, Dong \textit{et al.} \cite{dong2018few} proposed a prototype learning component in a framework of semantic segmentation that learned to take discriminative information from features to help segment objects better. Also, Wang \textit{et al.} \cite{Wang_2019_ICCV} introduced a prototype align method that learns class-specific prototype representations from a few image samples to perform segmentation over the query images. 
Lately, Liu \textit{et al.} proposed a dynamic prototype convolution network to address few-shot semantic segmentation. 
The work of \cite{Saha_2022_CVPR} proposed context-aware prototype learning. \cite{Tian_2022_CVPR} introduced generative models approach for this task.
Recently, Nguyen \textit{et al.} \cite{nguyen2022ifs} came up with iFS-RCNN, an instance segmenter via an incremental approach. Gao \textit{et al.} \cite{gao2022dc} proposed the DCFS framework, an effective decoupling classifier that boosted the performance of object detection and segmentation heads. Han \textit{et al.} \cite{han2023reference} suggested a reference twice transformer-based framework (RefT) to enhance features in segmentation tasks. Also in the transformer approach, Wang \textit{et al.} \cite{wang2022dynamic} introduced DTN to directly segment the target object instances from arbitrary categories given reference images. 

\hlc[white]{In common, these aforementioned methods of the two approaches including FSOD and FSS mostly focus on generic objects, which cannot create effective distinguished features and fail to recognize camouflaged objects. In our case, our proposed methods aim at highlighting the differences between backgrounds and foregrounds which we considered as the key feature to detect or segment camouflaged objects. Furthermore, our proposed approaches contribute directly to the training process of such models via loss functions.}

\begin{table}[!t]
\begin{center}
\begin{adjustbox}{max width=0.5\textwidth}
\begin{tabular}{|c|c|c|}
\hline
\#Instances & Ratio (\%)   & \#Images\\ \hline
1                          & 85.55 & 2440\\ \hline
2                          & 8.31 & 237 \\ \hline
3                          & 3.44 & 98\\ \hline
3$+$                    & 2.70 & 77\\ \hline
\end{tabular}
\end{adjustbox}
\end{center}
\caption{\hlc[white]{Number of instances per image of CAMO-FS.}}
\label{table:num_instance}
\end{table}

\begin{table*}[!h]
\caption{Extra collected number of images and instances in CAMO-FS dataset.}
\label{table:extra_data}
\begin{center}
\begin{adjustbox}{max width=\textwidth}
\begin{tabular}{|c|c|c|c|c|c|c|c|c|c|c|c|c|}
\hline
\textbf{Classes}    & Bat & Bear & Camel & Dolphin & Elephant & Horse & Kangaroo & Monkey & Penguin & Rhino & Squirrel & Total \\ \hline
\textbf{\#images }  & 12  & 14   & 14    & 13      & 14       & 16    & 22       & 16     & 11      & 14    & 17  &  163    \\ \hline
\textbf{\#instances} & 12  & 14   & 15    & 19      & 14       & 17    & 25       & 20     & 14      & 14    & 17   &  181  \\ \hline
\end{tabular}
\end{adjustbox}
\end{center}
\end{table*}

\begin{figure}[tb]
\begin{center}
  \includegraphics[width=0.5\textwidth]{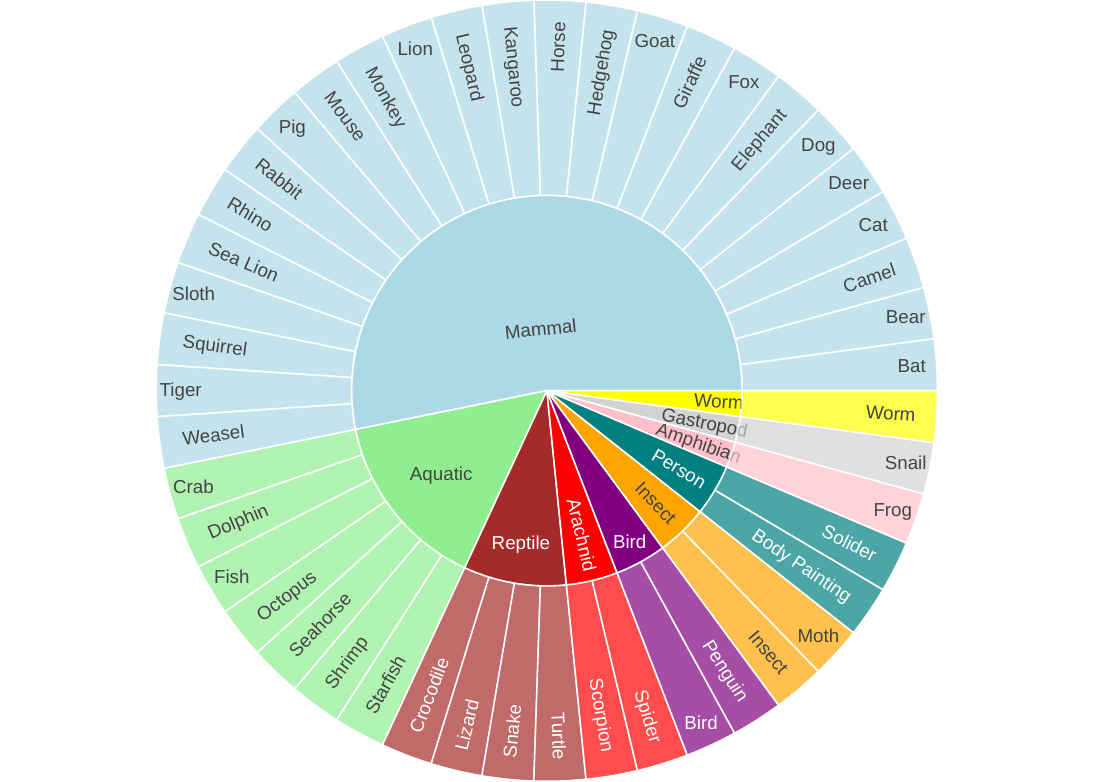}
\end{center}
  \caption{\hlc[white]{Hierarchical taxonomic structure of our CAMO-FS dataset.}}
\label{fig:vison taxonomic structure}
\end{figure}

\section{Proposed Method}
\label{proposed_method}

\subsection{CAMO-FS Benchmark Dataset}
\label{sec:camo-fs}
Camouflaged data tends to be more difficult to collect in the real world rather than non-camouflaged ones. Generating intensive annotations with multi-task or hierarchical labels for camouflaged objects is also costly and complicated, especially with the pixel level as polygon masks. Particularly, the visual characteristics of a camouflaged object are extremely identical to the background. The external appearances (i.g. the intensity, color, and textures) are close to their surrounding environment, the boundary between camouflaged objects and the background or other identical-type camouflaged objects in case of being nearly or partly overlapped. Thus, it is really tough to provide the concurrence between annotators due to ambiguity in verifying camouflaged regions blended in surroundings. For ease of data preparation such as collections and annotations, one of the most common ways is to inherit existing camouflaged datasets and CAMO++~\cite{le2021camouflaged} is our selected dataset since it is a high-diversity dataset with a variety of camouflaged object categories. Furthermore, the key to few-shot learning lies in the generalization ability of the pertinent model when presented with a few available samples. The context of camouflaged objects inherently matches this understanding because the number of camouflaged images is often scarce in practice.

\noindent\textbf{CAMO++ Dataset.} CAMO++ generally contains camouflaged and non-camouflaged images with a total of 5,500 images corresponding to 32,756 instances \cite{le2021camouflaged}. The dataset contains 93 fine-grained classes assigned to 13 coarse-grained classes. However, in the case of camouflaged objects, there are 47 fine-grained classes designed with a hierarchical structure and assigned into 10 coarse-grained classes. In detail, CAMO++ contributes 2,695 camouflage images including 1,250 existing camouflage images in the previous CAMO dataset with 1,450 newly collected camouflage images for CAMO++. In this scope of our paper, 2,800 remaining non-camouflage images are ignored. CAMO++ especially provides common ground truths such as bounding boxes, object masks, and instance masks which are suitable for many tasks of camouflage research.

\begin{figure}[t]
\begin{center}
  \includegraphics[width=\linewidth]{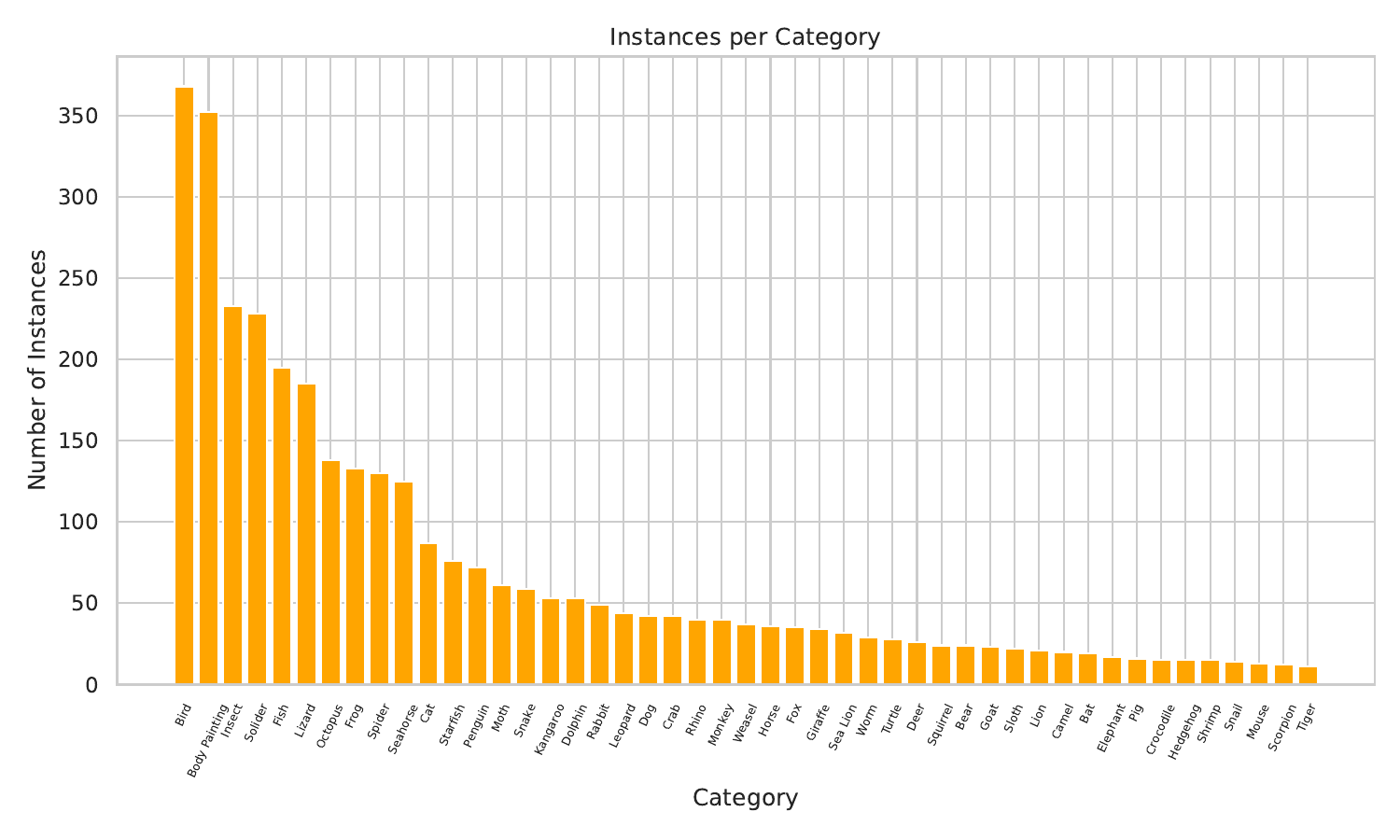}
\end{center}
  \caption{\hlc[white]{The distribution of CAMO-FS dataset. The categories are sorted.}}
\label{fig:camo-distribution}
\end{figure}

\noindent\textbf{CAMO-FS Dataset.} 
\hlc[white]{In general, there are three steps to construct the CAMO-FS dataset: \textbf{inheritance}, \textbf{collection}, \textbf{data splitting}. 

In the \textbf{inheritance} step,} we leverage the available CAMO++ to build our CAMO-FS dataset. In this way, we inherit the biology taxonomic and vision taxonomic structure of CAMO++ which helps us to reduce the burden of data collection.
{\color{black} \textcolor{red}{Table} \ref{table:camo_data_statistic} provides an overview of previous works done on camouflage, which is mentioned in the related work, and our proposed CAMO-FS in terms of main characteristics. We exploit the diversity of CAMO++ by its 10 meta-categories to build up the few-shot concept for instance segmentation. To this end, our CAMO-FS not only keeps a good ratio of instances per image of 1.172 but also contributes as the very first dataset specific for few-shot research on camouflaged animals. Note that the large amount of images in some datasets does not mean they are all camouflaged images.} 


\hlc[white]{In the \textbf{collection} step, as CAMO++ faces issues such as} imbalanced data and a shortage of the number of images of some classes, which cause evaluation problems for few-shot tasks. Particularly, there are 11 classes (\textit{e.g. Camel, Dolphin, Elephant, Horse, Kangaroo, Monkey, Penguin, Bat, Bear, Squirrel, and Rhino}) having a shortage of images that are needed to train a few-shot model. Hence, we hardly perform training or testing on these classes. As a result, we \hlc[white]{manually} collect more data for these classes with 163 total images corresponding to 181 instances (an average of 15-16 instances per class) \hlc[white]{on Google Image Search Engine with their class names as the search query}. We also remove images with mistakes in the original dataset. The statistics of collected data are shown in \textcolor{red}{Table} \ref{table:extra_data}. By gathering more camouflaged animals and combining them with the CAMO++ dataset, we conduct our CAMO-FS dataset for few-shot camouflaged animal detection and segmentation with $2,852$ total images corresponding to $3,342$ instances. \hlc[white]{\mbox{\textcolor{red}{Figure}~\ref{fig:vison taxonomic structure}} shows the vision taxonomic structure of coarse-grained and corresponding fine-grained classes and illustrates the ratios of 10 coarse-grained classes in our proposed CAMO-FS dataset. We also show the distribution of 47 camouflaged classes in \mbox{\textcolor{red}{Figure}~\ref{fig:camo-distribution}}, which indicates that CAMO-FS is a diverse and long-tailed dataset.} \textcolor{red}{Figure}~\mbox{\ref{fig:data_sample}} shows exemplary images with mask annotations from our proposed CAMO-FS.

\begin{figure}[!t]
    \centering
    \subfloat[\centering Our CAMO-FS]{{\includegraphics[width=4cm]{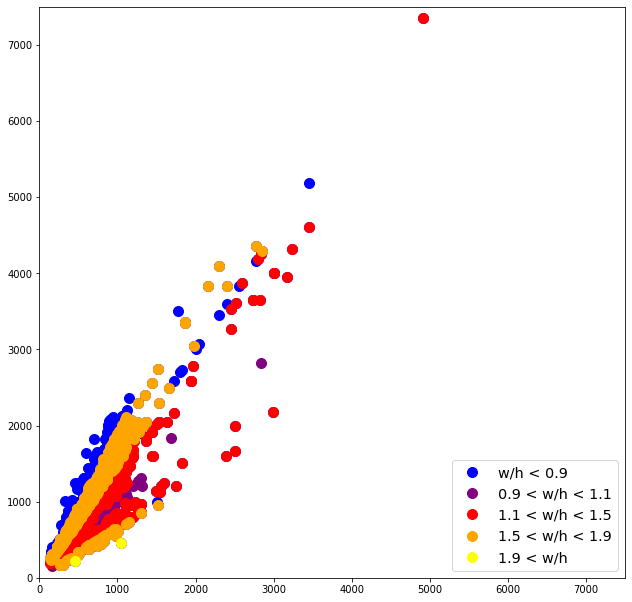} }}%
    \hspace{0.1em}
    \subfloat[\centering CAMO \cite{ltnghia-CVIU2019} ]{{\includegraphics[width=4cm]{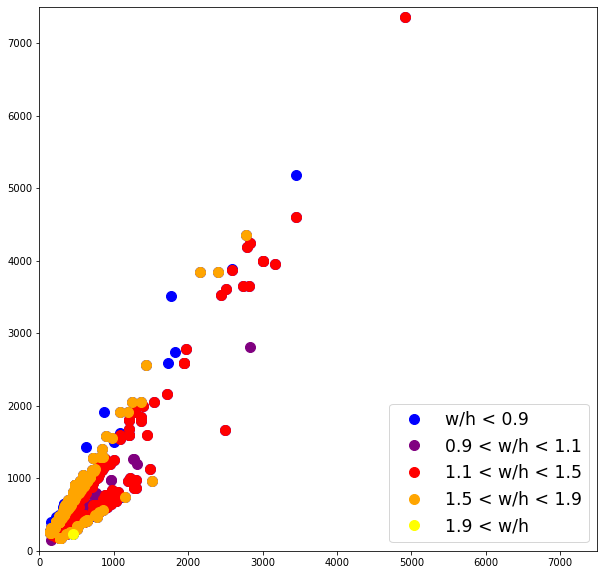} }}%
    \hspace{0.1em}
    \subfloat[\centering CAMO++ \cite{le2021camouflaged} ]{{\includegraphics[width=4cm]{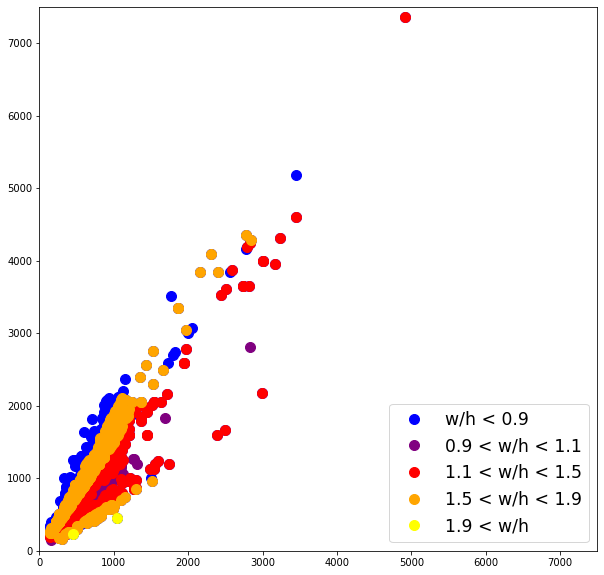} }}%
    \hspace{0.1em}
    \subfloat[\centering COD \cite{fan2020camouflaged} ]{{\includegraphics[width=4cm]{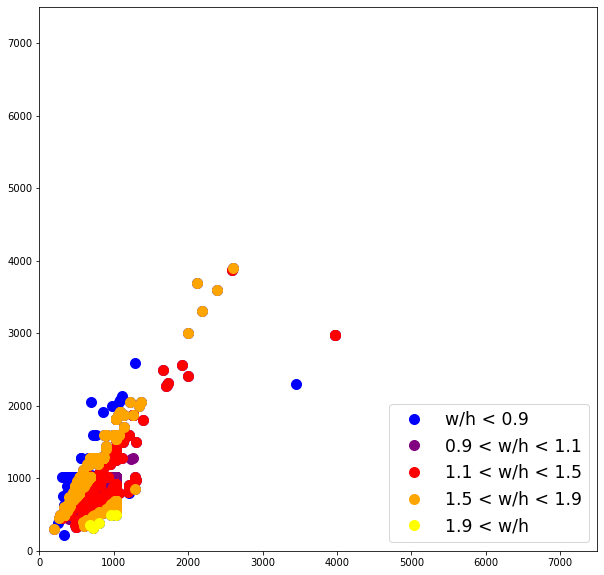} }}%
    \caption{{\color{black}Distribution of camouflage image resolution. Best viewed online in color and zoomed in.}}%
    \label{fig:reso}%
\end{figure}

\begin{figure}[!t]
    \centering
    \subfloat[\centering Our CAMO-FS]{{\includegraphics[width=4cm]{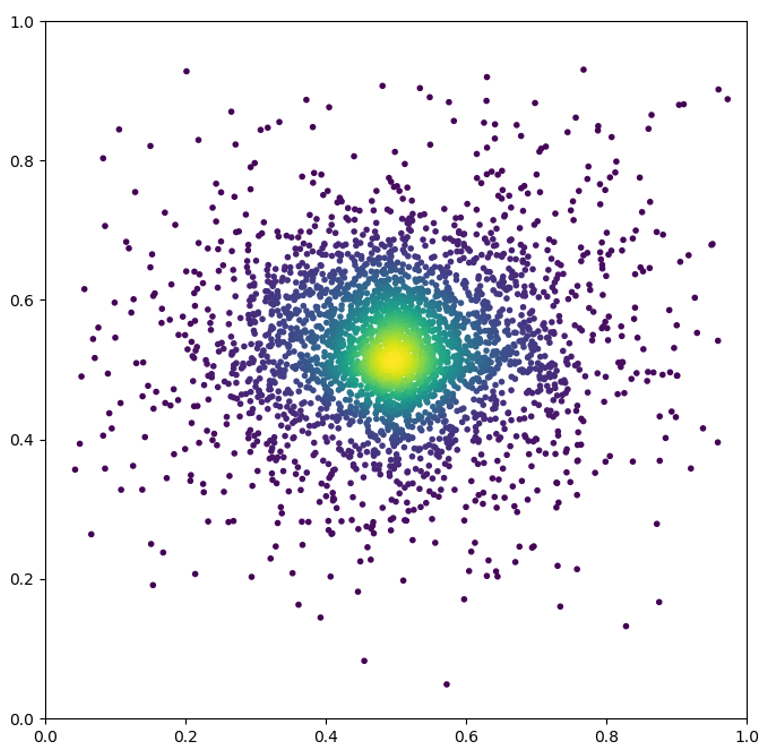} }}%
    \hspace{0.1em}
    \subfloat[\centering CAMO \cite{ltnghia-CVIU2019} ]{{\includegraphics[width=4cm]{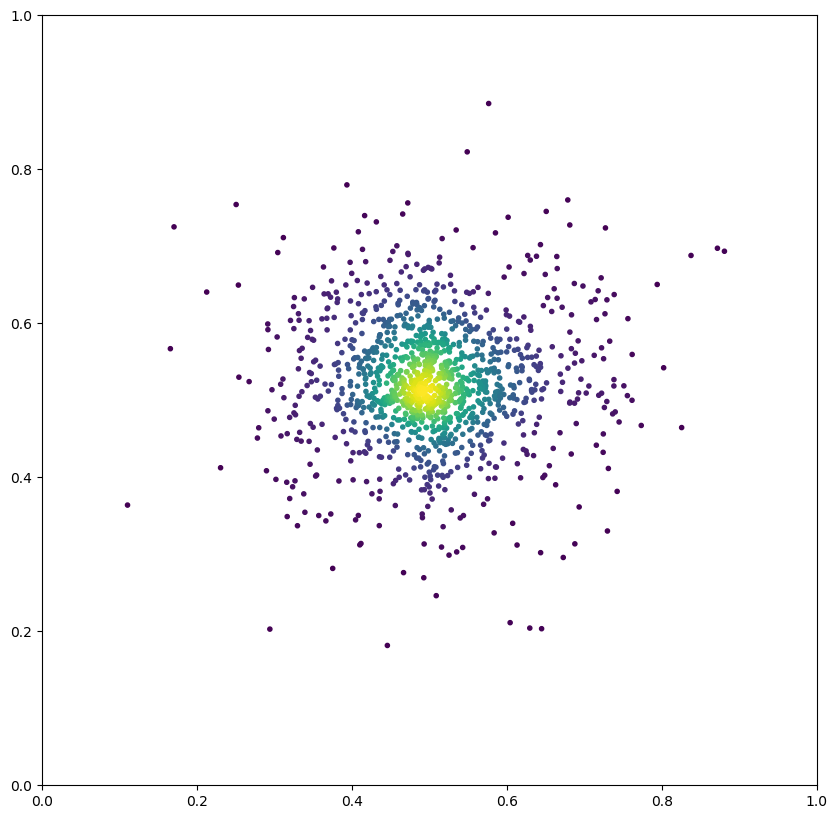} }}%
    \hspace{0.1em}
    \subfloat[\centering CAMO++ \cite{le2021camouflaged} ]{{\includegraphics[width=4cm]{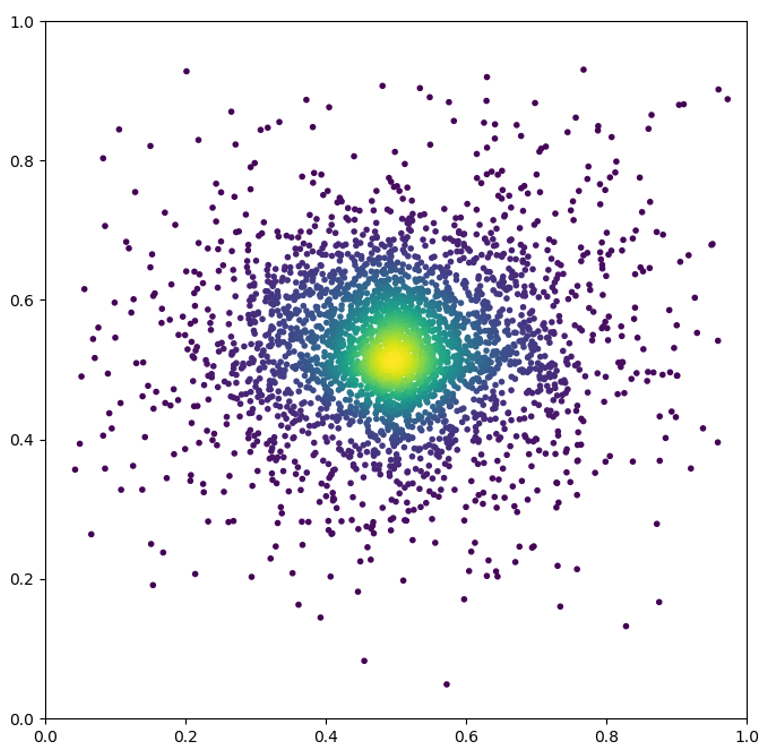} }}%
    \hspace{0.1em}
    \subfloat[\centering COD \cite{fan2020camouflaged} ]{{\includegraphics[width=4cm]{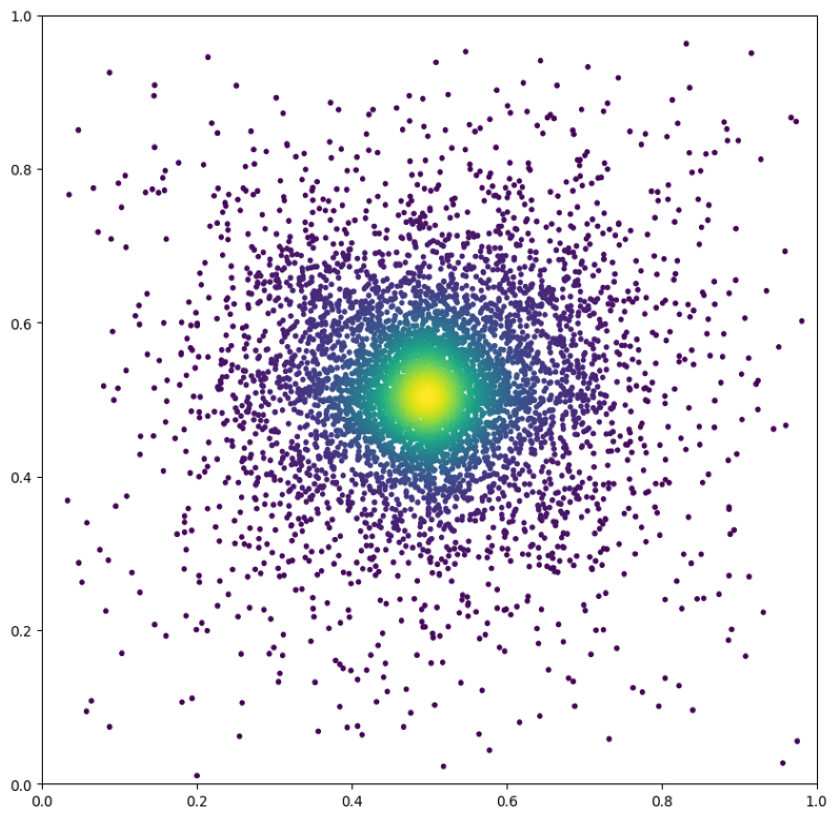} }}%
    \caption{{\color{black}Instance center bias camouflaged datasets. Best viewed online with color and zoomed in.}}%
    \label{fig:center_density}%
\end{figure}

{\color{black}In \textcolor{red}{Table} \ref{table:num_instance}, we report the aggregated number of instances per image. The number of instances per image ranges from 1 to 25 and commonly falls into 1, then 2 and 3 while the remaining is beyond 3 instances. As can be seen, the number of images that contain 1 to 3 instances takes up a large proportion of the entire dataset. This also illustrates the problem of data imbalance between the number of instances and the ratio of images in the dataset}, which reflects a problem that the presence of camouflaged animals captured in photos is often limited, i.e. mostly one animal per image. Additionally, although being claimed in~\cite{le2021camouflaged} that camouflaged objects in CAMO++ were localized over the entire image, after removing non-camouflage objects and adding new camouflaged images, we have the distributions of object centers in normalized image coordinates over all images in the CAMO-FS dataset as in \textcolor{red}{Figure} \ref{fig:center_density}-a. This means camouflaged animals tend to be located in the center of images. Indeed, to capture images of camouflaged animals in the wild, photographers need to carefully focus on the animals, which leads to the central layout of collected images. 
{\color{black} Also in \textcolor{red}{Figure} \ref{fig:center_density}, we illustrate the center bias of camouflaged images in other CAMO \cite{ltnghia-CVIU2019} and COD \cite{fan2020camouflaged} datasets for better visual comparison.}
{\color{black} In \textcolor{red}{Figure} \ref{fig:reso}, we present the image resolution among camouflage datasets. As we only consider camouflaged images of CAMO++ \cite{le2021camouflaged} and COD \cite{fan2020camouflaged}, the density of our CAMO-FS is slightly higher than CAMO++ as a result of our extra collection of images presented in \textcolor{red}{Table} \ref{table:extra_data}. In comparison with the previous COD \cite{fan2020camouflaged} and CAMO \cite{ltnghia-CVIU2019}, our CAMO-FS image resolution distribution is more satisfying in diversity.}

\begin{figure*}[!t]
\centering
    \includegraphics[width=1\textwidth]{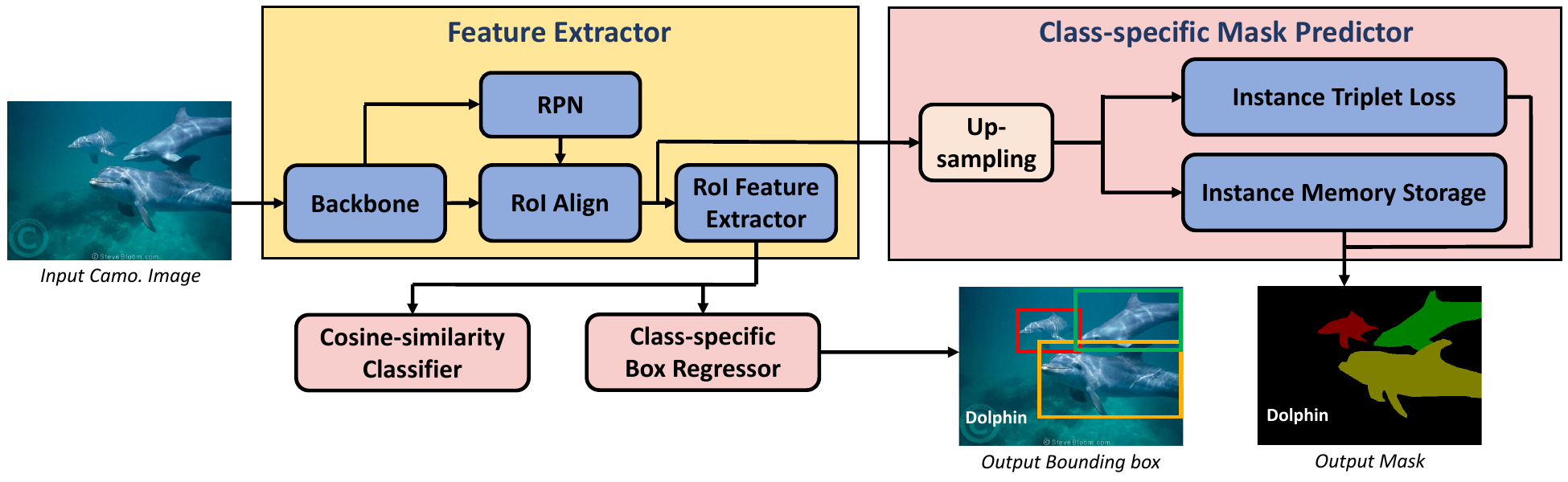}
\caption{Our general FS-CDIS framework for Few-Shot Camouflaged Detection and Instance Segmentation.}
\label{fig:general_fw}
\end{figure*} 

\hlc[white]{In the \textbf{data splitting} step}, to effectively create the data for the few-shot problem, we \hlc[white]{randomly get $M$ instances for each camouflaged class from the CAMO-FS dataset to create training sets. In our setup, $M$ equals $5$ for $47$ classes and thus there are $197$ training images containing $235$ camouflaged instances and the remaining $2,655$ images with $3,107$ instances are used for testing}. We only remove some objects of the higher-level training set if they exist to create the other few-shot settings. For example, we get all elements to generate 5-shot training data and discard 2 in 5 objects to make a 3-shot one. In this way, the 5-shot benchmark contains objects of the 3-shot dataset and the 3-shot setting contains the objects of the 2-shot one.

To the best of our knowledge, this is among the first works to address few-shot camouflaged instance segmentation and detection. Given the lack of a large-scale dataset for training and testing purposes on camouflaged animal issues, we build a benchmark for the task of few-shot camouflaged instance segmentation and detection. 

\subsection{General Framework}

\textbf{Few-shot instance segmentation formulation.}
In few-shot learning, we have one set of base classes denoted $C_{base}$ with a large amount of available training data, and one disjoint set of novel classes denoted $C_{novel}$ containing a small amount of training data. This amount is small to a few samples. The ultimate goal is to train a model to predict well on the novel classes $C_{test} = C_{novel}$ \cite{snell2017prototypical, vinyals2016matching} or on both base and novel data $C_{test} = C_{base} \cup C_{novel} $ \cite{gidaris2018dynamic}. 
In few-shot classification, this work \cite{vinyals2016matching} introduces the method of episodic training. The method sets up a series of episodes $E_i = (I_q, S_i)$ where $S_i$ is a support set that contains $N$ classes from $C_{train} = C_{novel} \cup C_{base}$ along with $K$ examples per class (so-called $N$-way $K$-shot). 
A network is then trained to classify an input image $I_q$, termed query image, out of the classes in $S_i$. The key idea is that solving a different classification task for each episode leads to better generalization and results on $C_{novel}$. The extended versions of this method are FSOD \cite{yolo-reweighting} and FSIS \cite{fan2020fgn, meta-rcnn}. Those proposals consider all objects in an image as queries and they have a single support set per image instead of per query.
However, there exist challenges in FSIS which are not only classification tasks but also how to determine their localization and segmentation. Use an image $I_q$ to query, FSIS returns labels $y_i$, bounding boxes $b_i$, and segmentation masks $M_i$ for all objects in $I_q$ that belong to the set of $C_{test}$. 


\noindent\textbf{General framework.}
Originated from TFA \cite{TFA} which uses Faster R-CNN \cite{Ren-NeurIPS2015}, MTFA \cite{Ganea_2021_CVPR} employs a mask prediction branch to return the pixel-wise mask for the segmentation task.
In this work, we leverage the architecture of MTFA model \cite{Ganea_2021_CVPR} based on Mask R-CNN \cite{Kaiming-ICCV2017} which is a two-stage training and fine-tuning mechanism. We train the first stage of the framework on 80 classes from the COCO dataset. This stage results in the base model weights for the second stage of novel fine-tuning. In the fine-tuning stage, we apply the few-shot technique to learn the novel concepts of camouflaged instances in our proposed CAMO-FS dataset.

Similar to Mask R-CNN, the input images are fed into a feature extractor $F$ consisting of backbone $B$, RoI Align, RoI feature extractor modules, and a region proposal network. There are three heads specifying three tasks that this scheme supports: a classification head $C$, a box regression head $R$, and a new attached mask prediction head $M$.
In the first stage, the network is trained on the base classes $C_{base}$. Then in the second stage, we froze the backbone network $B$ of the feature extractor $F$ and only perform training on the prediction heads. Thus, only RoI classifier $C$, box regressor $R$, and mask predictor $M$ are fine-tuned in the second stage.
In \textcolor{red}{Figure} \ref{fig:general_fw}, there exists a branch called mask predictor $M$. We apply similarly to Ganea \textit{et al.} \cite{Ganea_2021_CVPR} by using this two-stage fine-tuning approach. Firstly, the network is trained on base classes with lots of abundant data and then fine-tuning all predictor heads $C$, $R$, and $M$ on novel data of $K$ shots for each class.

Not a simple mask predictor $M$ that we use, we enhance the performance of the instance segmentation task by employing the two concepts of instance triplet loss and instance memory storage which are clearly described in the next section. The two improvements are inspired not only by the instance segmentation task in general but also by the camouflaged instance segmentation specifications.


\begin{figure*}[!t]
    \centering
    \includegraphics[width=\textwidth]{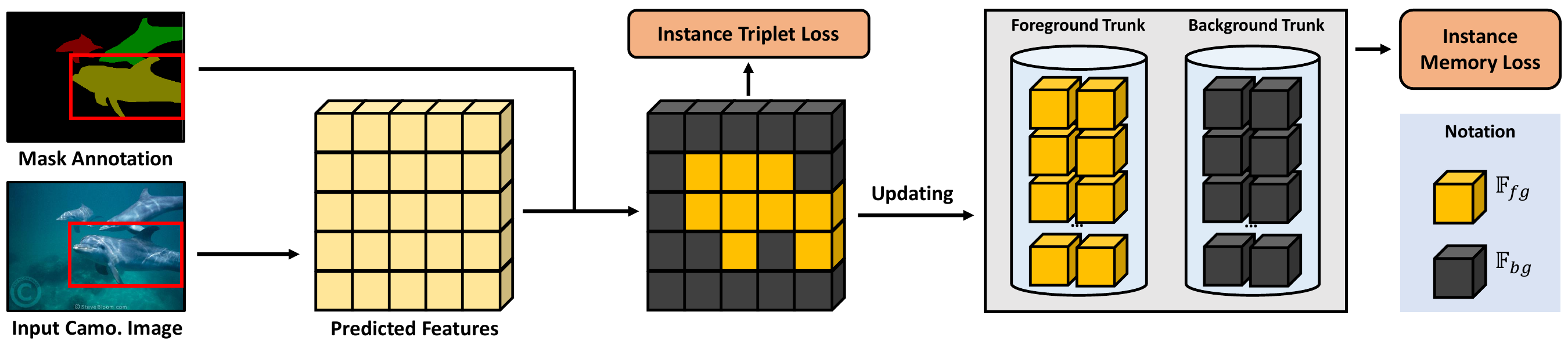}
    \caption{\hlc[white]{Visualization of instance triplet loss and instance memory loss for region proposal.}}
    \label{fig:framework_fs-cdis-memo-redesign-ieee-access}
\end{figure*}

\subsection{Framework Improvement}
One of the characteristics of camouflage instances is the camouflage texture similar to the background. This makes the precise identification of the boundary areas difficult. It is more critical in the context of few-shot learning where the concepts of a class are represented by only a few samples.

In this work, we thus propose improvements to enhance distinguishable features between background and foreground areas. In particular, we explore two approaches that focus on loss functions. The first one is the triplet loss function which was known as a strong metric to support the network in creating discrimination features between anchor and negative. The second approach is the idea of memory bank, which is used to enhance the distance between foreground and background not only for individual instances but also for each novel class. To this end, our framework is named after FS-CDIS.

To calculate the loss function, we employ the mask annotation for RoI features to collect the $\mathbb{F}_{bg}$ background and $\mathbb{F}_{fg}$ foreground features by location on each RoI. Both $\mathbb{F}_{bg}$ and $\mathbb{F}_{fg}$ for each proposal are used to calculate the respective loss functions which are presented in the following sections.

\subsubsection{Instance triplet loss}
{\color{black}With the idea of enhancing the discrimination between camouflaged instances and their backgrounds, we leverage the power of the triplet loss function \cite{balntas2016learning}. Specifically, we treat the pixels of an object as positive points and the background as negative ones. Accordingly, we force the model to learn the distinguished features among the foreground and background representatives. The more distinguished among features, the better a model can do to detect or segment camouflaged instances. In this way, we highlight the camouflaged instances so that the model is able to recognize them.}

For each RoI, we consider the average foreground features $\mathbb{F}_{avg} =\frac{1}{ | \mathbb{F}_{fg} | } \sum \mathbb{F}_{fg}$ 
as anchors with the foreground feature $\mathbb{F}_{fg}$ as positive and the background feature $\mathbb{F}_{bg}$ as negative to apply the triplet loss function \cite{balntas2016learning}. In this way, the model tries to learn to minimize the distance between foreground representatives and maximize the distance between background representatives as shown in \textcolor{red}{Figure} \ref{fig:framework_fs-cdis-memo-redesign-ieee-access}.  We use cosine similarity to calculate the distance instead of Euclidean distance. The loss function is defined as:

\begin{equation} \label{eq1}
\begin{split}
\small
\mathcal{L}_{triplet} & = \max\{d(\mathbb{F}_{avg}, \mathbb{F}_{fg}) - d(\mathbb{F}_{avg}, \mathbb{F}_{bg}) + margin, 0\} \\
 d(x,y) & = 1 - \frac{x{\cdot}y}{\|x\|{\cdot}\|y\|}
\end{split}
\end{equation}
, where $margin$ controls the discrimination between foreground and background features. In our experiments, we set $margin$ of 0.5. 

\subsubsection{Instance memory storage}
The memory bank is designed to store information within a class and the class information is updated during the training. Still, the model can learn information at a global level and has high consistency for each class. On the other hand, storing and updating the features in the memory bank for each iteration during training also creates more variants. By leveraging these advantages, we propose the memory bank for few-shot camouflage instance segmentation. To be specific, we use the memory bank to contain the background and foreground features per each class and make use of features to calculate the discrimination between areas of object and no object in region proposals (shown in \textcolor{red}{Figure} \ref{fig:framework_fs-cdis-memo-redesign-ieee-access}).

\begin{table*}[]
\caption{{\color{black} State-of-the-art comparison on CAMO-FS dataset among the baseline model of MTFA \cite{Ganea_2021_CVPR}, Mask RCNN$^\dagger$ \cite{Kaiming-ICCV2017}, iFS-RCNN \cite{nguyen2022ifs}, and our proposed methods FS-CDIS with instance triplet loss (ITL) and instance memory storage (IMS). Our performance improves over the utilized baselines.
}}
\label{table:sota_comparison}
\centering
\adjustbox{max width=\textwidth}{
\begin{tabular}{|lc|cccccccccc|}
\hline
\multicolumn{2}{|c|}{\textbf{Model}} & \multicolumn{10}{c|}{\textbf{Novel AP}} \\ \hline
\multicolumn{1}{|l|}{\multirow{2}{*}{\textbf{Method}}} 
& \multicolumn{1}{|c|}{\multirow{2}{*}{\textbf{Baseline}}}
& \multicolumn{5}{c|}{\textbf{Instance Segmentation}} 
& \multicolumn{5}{c|}{\textbf{Object Detection}} \\
\multicolumn{1}{|l|}{} &  & \textbf{1} & \textbf{2} & \textbf{3} & \textbf{5} & \multicolumn{1}{c|}{\textbf{Avg.}} & \textbf{1} & \textbf{2} & \textbf{3} & \textbf{5} & \textbf{Avg.} \\ \hline
\multicolumn{1}{|l|}{\textbf{MTFA \cite{Ganea_2021_CVPR}}} & \multicolumn{1}{c|}{\multirow{3}{*}{\begin{tabular}[c]{@{}c@{}}COCO-80 ResNet-50\end{tabular}}} & 2.48 & 6.67 & 5.81 & 6.40 & \multicolumn{1}{c|}{5.34} & 1.98 & 6.47 & 5.82 & 6.17 & 5.11 \\
\multicolumn{1}{|l|}{\textbf{M-RCNN$^\dagger$ \cite{Kaiming-ICCV2017}}} &  & 4.08 & 6.79 & 6.90 & 8.29 & \multicolumn{1}{c|}{6.52} & 2.82 & 5.09 & 5.46 & 6.18 & 4.89 \\
\multicolumn{1}{|l|}{\textbf{iFS-RCNN \cite{nguyen2022ifs}}} &  & 4.17 & 6.26 & 5.73 & 6.38 & \multicolumn{1}{c|}{5.64} & 3.92 & 6.06 & 5.47 & 6.60 & 5.51 \\ \hline
\multicolumn{1}{|l|}{\textbf{MTFA \cite{Ganea_2021_CVPR}}} & \multirow{3}{*}{\begin{tabular}[c]{@{}c@{}}COCO-80 ResNet-101\end{tabular}} & 3.66 & 6.21 & 6.16 & 5.95 & \multicolumn{1}{c|}{5.50} & 2.93 & 5.90 & 5.84 & 5.84 & 5.13 \\
\multicolumn{1}{|l|}{\textbf{M-RCNN$^\dagger$ \cite{Kaiming-ICCV2017}}} &  & 4.39 & 7.69 & 7.94 & 10.09 & \multicolumn{1}{c|}{7.53} & 3.03 & 5.80 & 6.20 & 7.79 & 5.71 \\
\multicolumn{1}{|l|}{\textbf{iFS-RCNN \cite{nguyen2022ifs}}} &  & 4.27 & 6.55 & 6.07 & 7.80 & \multicolumn{1}{c|}{6.17} & 3.79 & 6.28 & 6.01 & 8.08 & 6.04 \\ \hline
\multicolumn{12}{|c|}{\textbf{Our performance }} \\ \hline
\rowcolor{Gray}
\multicolumn{1}{|l|}{\textbf{FS-CDIS-ITL}} & 
& 4.46 & 5.57 & 6.41 & 8.48 & \multicolumn{1}{c|}{\textbf{6.23}} & 4.04 & 7.28 & 7.49 & 9.76 & \textbf{7.14} \\
\rowcolor{Gray}
\multicolumn{1}{|l|}{\textbf{FS-CDIS-IMS}} & \raisebox{1ex}[1ex]{ResNet-101 MTFA} & 5.46 & 6.95 & 7.36 & 9.61 & \multicolumn{1}{c|}{\textbf{7.35}} & 4.50 & 6.95 & 7.55 & 10.36 & \textbf{7.34} \\ \hline
\rowcolor{Gray}
\multicolumn{1}{|l|}{\textbf{FS-CDIS-ITL}} &  & 5.73 & 7.97 & 8.52 & 9.92 & \multicolumn{1}{c|}{\textbf{8.04}} & 5.08 & 7.56 & 7.85 & 9.67 & \textbf{7.34} \\
\rowcolor{Gray}
\multicolumn{1}{|l|}{\textbf{FS-CDIS-IMS}} & \raisebox{1ex}[1ex]{ResNet-101 M-RCNN} & 5.52 & 7.84 & 8.65 & 9.82 & \multicolumn{1}{c|}{\textbf{7.96}} & 4.92 & 7.39 & 7.96 & 9.52 & \textbf{7.45} \\ \hline
\rowcolor{Gray}
\multicolumn{1}{|l|}{\textbf{FS-CDIS-ITL}} &  & 5.35 & 6.01 & 7.80 & 9.35 & \multicolumn{1}{c|}{\textbf{7.13}} & 4.71 & 5.66 & 7.10 & 10.36 & \textbf{6.96} \\
\rowcolor{Gray}
\multicolumn{1}{|l|}{\textbf{FS-CDIS-IMS}} & \raisebox{1ex}[1ex]{ResNet-101 iFS-RCNN}& 2.99 & 6.83 & 6.14 & 9.03 & \multicolumn{1}{c|}{\textbf{6.25}} & 2.74 & 6.39 & 5.94 & 8.44 & \textbf{5.88} \\ \hline
\multicolumn{12}{l}{\multirow{1}{*}{\textit{\begin{tabular}[c]{@{}l@{}} M-RCNN$^\dagger$ is Mask R-CNN \cite{Kaiming-ICCV2017} with sigmoid classifier.\end{tabular}}}}
\end{tabular}}
\end{table*}

\noindent \textbf{Storing and updating}: The memory bank for each class contains 2N features including N of foreground features and N of background features. While the memory bank receives new features, the module concatenates them with existing old features. In case the number of features is greater than the given N features, the memory bank releases the oldest features to maintain the number of features to N. This process updates the features in the memory bank and keeps the quantity of the stored features appropriate to the memory size (also known as the memory capacity).

\noindent \textbf{Sampling}: To calculate the loss value, the memory bank has to provide three elements $\mathbb{F}_{fg}$, $\mathbb{F}_{bg}$, and $\mathbb{F}_{general}$. $\mathbb{F}_{fg}$ and $\mathbb{F}_{bg}$ are all foreground and background features that module storing. The $\mathbb{F}_{general}$  is the general foreground feature, and it is created for each class by averaging the $\mathbb{F}_{fg}$.

Let $\mathbb{F}_{fg}^i$ be the i-th foreground feature and $\tau$ be a temperature hyper-parameter in \cite{wu2018unsupervised}. In our experiments, we set $\tau$ as 1. 
The memory loss function for camouflaged instances is introduced as follows:
\begin{equation}
    \small
    \mathcal{L}_{memory} = -\log\frac{\exp(\mathbb{F}_{general}{\cdot}\mathbb{F}_{fg}^i / \tau)}{\sum_{j=0}^{|\mathbb{F}_{bg}|}\exp(\mathbb{F}_{general}{\cdot}\mathbb{F}_{bg}^j  / \tau) +  \exp(\mathbb{F}_{general}{\cdot}\mathbb{F}_{fg}^i / \tau)}
    \label{eq:infonce}
\end{equation}

To this end, the final loss of our training process, which contains an instance triplet loss and memory storage is defined as follows:

\begin{equation} \label{eq4}
\mathcal{L}_{final} = \mathcal{L}_{mrcnn} + \alpha \mathcal{L}_{triplet} + \beta \mathcal{L}_{memory}. 
\end{equation}
Here, the parameter $\alpha$ of $\mathcal{L}_{triplet}$ and $\beta$ of $\mathcal{L}_{memory}$ are used during the training process to keep the balance between the two loss functions. Details of these functions are mentioned in the following section.

\section{Experiments}
\label{experiments}

We first overview the metrics and the experiment settings and the implementation details in \textcolor{red}{Section} \ref{overview_experiment} and then we evaluate and discuss our improvement on the general framework, {\color{black} as well as ablation study for our core proposed methods} in \textcolor{red}{Section} \ref{result_experiment}. 

\subsection{Overview}
\label{overview_experiment}
As specified in this work, we utilize the proposed CAMO-FS dataset containing images of camouflaged animals in the wild to establish the evaluation of our baseline and proposed improvement. We follow the concept procedure published in FSOD \cite{TFA, yolo-reweighting, meta-rcnn}. \hlc[white]{We employed the MTFA baseline~\mbox{\cite{Ganea_2021_CVPR}} implemented using Detectron2 framework \mbox{\cite{detectron2}}. The backbone is ResNet-101 \mbox{\cite{he2016deep}} with Feature Pyramid Network \mbox{\cite{Lin-CVPR2017}}. The models are trained in two stages: base training and novel fine-tuning stage.}

In the first stage of the base phase, we train our model with abundant data from $80$ classes \hlc[white]{with $118K$ images in the \texttt{train2017} set of the COCO dataset. The training hyper-parameters of the base phase are set according to Detectron2 settings~\mbox{\cite{detectron2}}}. 

\begin{table}[bt]
\centering
\caption{\hlc[white]{The number of instances for each camouflaged class in the test set of our CAMO-FS.}}
\label{table:test_set}
\adjustbox{max width=\linewidth}{
\begin{tabular}{|ll|ll|ll|}
\hline
\textbf{Class} & \textbf{\#ins.} & \textbf{Class} & \textbf{\#ins.} & \textbf{Class} & \textbf{\#ins.} \\\hline
Shrimp & 10 & Crab & 37 & Dolphin & 48 \\
Crocodile & 10 & Snake & 54 & Turtle & 23 \\
Worm & 24 & Snail & 9 & Kangaroo & 48 \\
Elephant & 12 & Giraffe & 29 & Goat & 18 \\
Leopard & 39 & Monkey & 35 & Deer & 21 \\
Sea\_Lion & 27 & Lion & 16 & Bat & 13 \\
Fox & 30 & Camel & 15 & Weasel & 32 \\
Rabbit & 44 & Dog & 37 & Horse & 31 \\
Mouse & 8 & Hedgehog & 10 & Rhino & 35 \\
Tiger & 6 & Pig & 11 & Squirrel & 19 \\
Sloth & 17 & Scorpion & 7 & Bear & 19 \\
Octopus & 133 & Starfish & 71 & Seahorse & 120 \\
Fish & 190 & Frog & 128 & Lizard & 180 \\
Cat & 82 & Moth & 56 & Penguin & 67 \\
Spider & 125 & Insect & 228 & Bird & 363 \\\cline{5-6}
Body\_Paint. & 347 & Solider & 223 & \textbf{Total} & \textbf{3107}\\\hline
\end{tabular}
}
\end{table}

In the second stage of the fine-tuning phase, we evaluate the performance of having $K = \{1, 2, 3, 5\}$ shots per each novel class. \hlc[white]{Specifically, in the 5-shot setting, we train the novel detector on $47$ camouflaged classes with $197$ images of the CAMO-FS dataset. The training set for other settings is a subset of the 5-shot setting (as presented in~\mbox{\textcolor{red}{Section}~\ref{sec:camo-fs}}). The novel phase has a learning rate of $0.00125$ inferred from the MTFA configuration. We set the balance parameters $\alpha$ = $1 \times 10^{-1}$ and $\beta$ = $1 \times 10^{-2}$ when we train the model with instance triplet loss and instance memory storage, respectively. Other training hyper-parameters of the novel phase are set following TFA~\mbox{\cite{TFA}} settings. Then, the novel models are assessed in a test set including $2,655$ images with $3,107$ instances of $47$ camouflaged classes to obtain results. The number of instances for each class is detailed in~\mbox{\red{Table \ref{table:test_set}}}.
Please visit ~\mbox{\cite{TFA}} or~\mbox{\cite{detectron2}} for more details on other parameters of both the training and testing phases. Our models are trained and tested on a single GeForce RTX 2080 Ti GPU. The frame per second (FPS) is approximately 15.}

To report our results on detection and instance segmentation tasks, we use average precision (AP) and average recall (AR). To be detailed, we report AP@50 and AP@75, along with AR@10. Besides, we also report AP and AR at small, medium, and large scales of the instances to further understand the model performance. {\color{black} For more details, readers can visit the homepage of the COCO dataset for detection and segmentation evaluation metrics \footnote{https://cocodataset.org/\#detection-eval}.}

\begin{table*}[!t]
\caption{\hlc[white]{The improvement of our proposed instance triplet loss (ITL) and instance memory storage (IMS) over the baseline MTFA} \cite{TFA}. The best performances are marked in \textcolor{black}{\textbf{bold}}. \# denotes the Number of shots.}
\label{table:improvement_result}
\centering
\adjustbox{width=1\textwidth}{
\begin{tabular}{|clrrrrrrrrrrr|}
\hline
\multicolumn{1}{|c|}{\textbf{\#}} & \multicolumn{1}{c|}{\textbf{Method}} & \multicolumn{1}{c|}{\textbf{AP}} & \multicolumn{1}{c|}{\textbf{AP50}} & \multicolumn{1}{c|}{\textbf{AP75}} & \multicolumn{1}{c|}{\textbf{APs}} & \multicolumn{1}{c|}{\textbf{APm}} & \multicolumn{1}{c|}{\textbf{APl}} & \multicolumn{1}{c|}{\textbf{AR1}} & \multicolumn{1}{c|}{\textbf{AR10}} & \multicolumn{1}{c|}{\textbf{ARs}} & \multicolumn{1}{c|}{\textbf{ARm}} & \multicolumn{1}{c|}{\textbf{ARl}} \\ \hline
\multicolumn{13}{|c|}{\textbf{Instance Segmentation}} \\ \hline
\multicolumn{1}{|c|}{} & \multicolumn{1}{l|}{\textbf{Baseline MTFA}} & \multicolumn{1}{r|}{3.66} & \multicolumn{1}{r|}{5.37} & \multicolumn{1}{r|}{4.09} & \multicolumn{1}{r|}{22.42} & \multicolumn{1}{r|}{4.35} & \multicolumn{1}{r|}{2.01} & \multicolumn{1}{r|}{11.30} & \multicolumn{1}{r|}{13.58} & \multicolumn{1}{r|}{25.97} & \multicolumn{1}{r|}{\textcolor{black}{ \textbf{12.96}}} & 12.53 \\
\multicolumn{1}{|c|}{} & \multicolumn{1}{l|}{\textbf{MTFA + ITL}} & \multicolumn{1}{r|}{4.46} & \multicolumn{1}{r|}{8.21} & \multicolumn{1}{r|}{4.60} & \multicolumn{1}{r|}{21.33} & \multicolumn{1}{r|}{4.13} & \multicolumn{1}{r|}{\textcolor{black}{ \textbf{4.01}}} & \multicolumn{1}{r|}{12.36} & \multicolumn{1}{r|}{15.04} & \multicolumn{1}{r|}{23.17} & \multicolumn{1}{r|}{9.49} & 16.67 \\
\multicolumn{1}{|c|}{} & \multicolumn{1}{l|}{\textbf{MTFA + IMS}} & \multicolumn{1}{r|}{\textcolor{black}{ \textbf{5.46}}} & \multicolumn{1}{r|}{\textcolor{black}{ \textbf{9.20}}} & \multicolumn{1}{r|}{\textcolor{black}{ \textbf{6.17}}} & \multicolumn{1}{r|}{\textcolor{black}{ \textbf{27.79}}} & \multicolumn{1}{r|}{\textcolor{black}{ \textbf{6.20}}} & \multicolumn{1}{r|}{\textcolor{black}{ \textbf{4.01}}} & \multicolumn{1}{r|}{{{17.08}}} & \multicolumn{1}{r|}{{{19.99}}} & \multicolumn{1}{r|}{\textcolor{black}{ \textbf{29.41}}} & \multicolumn{1}{r|}{11.45} & {{20.89}} \\
\multicolumn{1}{|c|}{\multirow{-4}{*}{\textbf{1}}} & \multicolumn{1}{l|}{\textbf{MTFA + Both}} & \multicolumn{1}{r|}{{{5.02}}} & \multicolumn{1}{r|}{{{8.58}}} & \multicolumn{1}{r|}{{{5.38}}} & \multicolumn{1}{r|}{{{26.29}}} & \multicolumn{1}{r|}{{{5.32}}} & \multicolumn{1}{r|}{{{3.93}}} & \multicolumn{1}{r|}{\textcolor{black}{ \textbf{17.98}}} & \multicolumn{1}{r|}{\textcolor{black}{ \textbf{21.42}}} & \multicolumn{1}{r|}{{{28.13}}} & \multicolumn{1}{r|}{{{12.68}}} & \textcolor{black}{ \textbf{22.64}} \\ \hline
\multicolumn{1}{|c|}{} & \multicolumn{1}{l|}{\textbf{Baseline MTFA}} & \multicolumn{1}{r|}{6.21} & \multicolumn{1}{r|}{8.92} & \multicolumn{1}{r|}{7.28} & \multicolumn{1}{r|}{32.64} & \multicolumn{1}{r|}{\textcolor{black}{ \textbf{7.75}}} & \multicolumn{1}{r|}{3.50} & \multicolumn{1}{r|}{18.88} & \multicolumn{1}{r|}{21.12} & \multicolumn{1}{r|}{\textcolor{black}{ \textbf{35.82}}} & \multicolumn{1}{r|}{\textcolor{black}{ \textbf{15.49}}} & 20.14 \\
\multicolumn{1}{|c|}{} & \multicolumn{1}{l|}{\textbf{MTFA + ITL}} & \multicolumn{1}{r|}{5.57} & \multicolumn{1}{r|}{9.45} & \multicolumn{1}{r|}{6.04} & \multicolumn{1}{r|}{25.83} & \multicolumn{1}{r|}{3.01} & \multicolumn{1}{r|}{5.37} & \multicolumn{1}{r|}{15.67} & \multicolumn{1}{r|}{17.33} & \multicolumn{1}{r|}{26.13} & \multicolumn{1}{r|}{7.37} & 17.50 \\
\multicolumn{1}{|c|}{} & \multicolumn{1}{l|}{\textbf{MTFA + IMS}} & \multicolumn{1}{r|}{{{6.95}}} & \multicolumn{1}{r|}{{{10.72}}} & \multicolumn{1}{r|}{{{7.60}}} & \multicolumn{1}{r|}{\textcolor{black}{ \textbf{33.62}}} & \multicolumn{1}{r|}{5.73} & \multicolumn{1}{r|}{{{6.44}}} & \multicolumn{1}{r|}{{{20.00}}} & \multicolumn{1}{r|}{{{22.15}}} & \multicolumn{1}{r|}{{{34.25}}} & \multicolumn{1}{r|}{13.86} & {{20.92}} \\
\multicolumn{1}{|c|}{\multirow{-4}{*}{\textbf{2}}} & \multicolumn{1}{l|}{\textbf{MTFA + Both}} & \multicolumn{1}{r|}{\textcolor{black}{ \textbf{7.60}}} & \multicolumn{1}{r|}{\textcolor{black}{ \textbf{11.37}}} & \multicolumn{1}{r|}{\textcolor{black}{ \textbf{8.26}}} & \multicolumn{1}{r|}{{{33.58}}} & \multicolumn{1}{r|}{{{5.97}}} & \multicolumn{1}{r|}{\textcolor{black}{ \textbf{6.49}}} & \multicolumn{1}{r|}{\textcolor{black}{ \textbf{22.76}}} & \multicolumn{1}{r|}{\textcolor{black}{ \textbf{24.85}}} & \multicolumn{1}{r|}{34.02} & \multicolumn{1}{r|}{{{15.01}}} & \textcolor{black}{ \textbf{24.50}} \\ \hline
\multicolumn{1}{|c|}{} & \multicolumn{1}{l|}{\textbf{Baseline MTFA}} & \multicolumn{1}{r|}{6.16} & \multicolumn{1}{r|}{8.95} & \multicolumn{1}{r|}{6.68} & \multicolumn{1}{r|}{33.74} & \multicolumn{1}{r|}{{{6.19}}} & \multicolumn{1}{r|}{5.08} & \multicolumn{1}{r|}{20.25} & \multicolumn{1}{r|}{22.95} & \multicolumn{1}{r|}{36.83} & \multicolumn{1}{r|}{16.31} & 21.63 \\
\multicolumn{1}{|c|}{} & \multicolumn{1}{l|}{\textbf{MTFA + ITL}} & \multicolumn{1}{r|}{6.41} & \multicolumn{1}{r|}{10.67} & \multicolumn{1}{r|}{6.72} & \multicolumn{1}{r|}{30.39} & \multicolumn{1}{r|}{5.17} & \multicolumn{1}{r|}{5.30} & \multicolumn{1}{r|}{20.69} & \multicolumn{1}{r|}{22.98} & \multicolumn{1}{r|}{31.90} & \multicolumn{1}{r|}{15.69} & 22.53 \\
\multicolumn{1}{|c|}{} & \multicolumn{1}{l|}{\textbf{MTFA + IMS}} & \multicolumn{1}{r|}{{{7.36}}} & \multicolumn{1}{r|}{{{11.23}}} & \multicolumn{1}{r|}{{{8.49}}} & \multicolumn{1}{r|}{{{37.03}}} & \multicolumn{1}{r|}{\textcolor{black}{ \textbf{6.24}}} & \multicolumn{1}{r|}{{{5.64}}} & \multicolumn{1}{r|}{\textcolor{black}{ \textbf{24.40}}} & \multicolumn{1}{r|}{\textcolor{black}{ \textbf{27.69}}} & \multicolumn{1}{r|}{{{38.44}}} & \multicolumn{1}{r|}{{{17.02}}} & \textcolor{black}{ \textbf{26.71}} \\
\multicolumn{1}{|c|}{\multirow{-4}{*}{\textbf{3}}} & \multicolumn{1}{l|}{\textbf{MTFA + Both}} & \multicolumn{1}{r|}{\textcolor{black}{ \textbf{7.85}}} & \multicolumn{1}{r|}{\textcolor{black}{ \textbf{11.74}}} & \multicolumn{1}{r|}{\textcolor{black}{ \textbf{9.05}}} & \multicolumn{1}{r|}{\textcolor{black}{ \textbf{37.57}}} & \multicolumn{1}{r|}{5.49} & \multicolumn{1}{r|}{\textcolor{black}{ \textbf{6.69}}} & \multicolumn{1}{r|}{{{24.33}}} & \multicolumn{1}{r|}{{{27.42}}} & \multicolumn{1}{r|}{\textcolor{black}{ \textbf{39.05}}} & \multicolumn{1}{r|}{\textcolor{black}{ \textbf{17.36}}} & {{25.92}} \\ \hline
\multicolumn{1}{|c|}{} & \multicolumn{1}{l|}{\textbf{Baseline MTFA}} & \multicolumn{1}{r|}{5.95} & \multicolumn{1}{r|}{8.67} & \multicolumn{1}{r|}{6.94} & \multicolumn{1}{r|}{34.71} & \multicolumn{1}{r|}{\textcolor{black}{ \textbf{6.25}}} & \multicolumn{1}{r|}{4.85} & \multicolumn{1}{r|}{21.29} & \multicolumn{1}{r|}{24.42} & \multicolumn{1}{r|}{36.86} & \multicolumn{1}{r|}{\textcolor{black}{ \textbf{14.51}}} & 24.83 \\
\multicolumn{1}{|c|}{} & \multicolumn{1}{l|}{\textbf{MTFA + ITL}} & \multicolumn{1}{r|}{{{8.48}}} & \multicolumn{1}{r|}{{{13.43}}} & \multicolumn{1}{r|}{{{9.80}}} & \multicolumn{1}{r|}{36.66} & \multicolumn{1}{r|}{5.75} & \multicolumn{1}{r|}{{{8.04}}} & \multicolumn{1}{r|}{23.83} & \multicolumn{1}{r|}{26.66} & \multicolumn{1}{r|}{37.03} & \multicolumn{1}{r|}{11.62} & 25.91 \\
\multicolumn{1}{|c|}{} & \multicolumn{1}{l|}{\textbf{MTFA + IMS}} & \multicolumn{1}{r|}{\textcolor{black}{ \textbf{9.61}}} & \multicolumn{1}{r|}{\textcolor{black}{ \textbf{14.61}}} & \multicolumn{1}{r|}{\textcolor{black}{ \textbf{11.73}}} & \multicolumn{1}{r|}{\textcolor{black}{ \textbf{38.60}}} & \multicolumn{1}{r|}{{{5.79}}} & \multicolumn{1}{r|}{\textcolor{black}{ \textbf{10.40}}} & \multicolumn{1}{r|}{\textcolor{black}{ \textbf{26.65}}} & \multicolumn{1}{r|}{\textcolor{black}{ \textbf{30.37}}} & \multicolumn{1}{r|}{{{39.21}}} & \multicolumn{1}{r|}{12.26} & \textcolor{black}{ \textbf{30.02}} \\
\multicolumn{1}{|c|}{\multirow{-4}{*}{\textbf{5}}} & \multicolumn{1}{l|}{\textbf{MTFA + Both}} & \multicolumn{1}{r|}{8.37} & \multicolumn{1}{r|}{13.29} & \multicolumn{1}{r|}{9.45} & \multicolumn{1}{r|}{{{38.44}}} & \multicolumn{1}{r|}{5.72} & \multicolumn{1}{r|}{7.60} & \multicolumn{1}{r|}{{{25.27}}} & \multicolumn{1}{r|}{{{29.31}}} & \multicolumn{1}{r|}{\textcolor{black}{ \textbf{39.33}}} & \multicolumn{1}{r|}{{{14.19}}} & {{28.52}} \\ \hline
\multicolumn{13}{|c|}{\textbf{Object Detection}} \\ \hline
\multicolumn{1}{|c|}{} & \multicolumn{1}{l|}{\textbf{Baseline MTFA}} & \multicolumn{1}{r|}{2.93} & \multicolumn{1}{r|}{5.86} & \multicolumn{1}{r|}{2.20} & \multicolumn{1}{r|}{20.95} & \multicolumn{1}{r|}{4.18} & \multicolumn{1}{r|}{2.03} & \multicolumn{1}{r|}{9.25} & \multicolumn{1}{r|}{10.84} & \multicolumn{1}{r|}{21.74} & \multicolumn{1}{r|}{{{11.49}}} & 8.77 \\
\multicolumn{1}{|c|}{} & \multicolumn{1}{l|}{\textbf{MTFA + ITL}} & \multicolumn{1}{r|}{4.04} & \multicolumn{1}{r|}{{{8.65}}} & \multicolumn{1}{r|}{2.98} & \multicolumn{1}{r|}{20.50} & \multicolumn{1}{r|}{4.90} & \multicolumn{1}{r|}{\textcolor{black}{ \textbf{4.22}}} & \multicolumn{1}{r|}{12.89} & \multicolumn{1}{r|}{{{15.53}}} & \multicolumn{1}{r|}{20.73} & \multicolumn{1}{r|}{11.45} & {{17.46}} \\
\multicolumn{1}{|c|}{} & \multicolumn{1}{l|}{\textbf{MTFA + IMS}} & \multicolumn{1}{r|}{\textcolor{black}{ \textbf{4.50}}} & \multicolumn{1}{r|}{\textcolor{black}{ \textbf{9.14}}} & \multicolumn{1}{r|}{{{3.45}}} & \multicolumn{1}{r|}{\textcolor{black}{ \textbf{22.88}}} & \multicolumn{1}{r|}{{{5.61}}} & \multicolumn{1}{r|}{3.54} & \multicolumn{1}{r|}{{{13.14}}} & \multicolumn{1}{r|}{15.22} & \multicolumn{1}{r|}{\textcolor{black}{ \textbf{23.14}}} & \multicolumn{1}{r|}{8.78} & 16.33 \\
\multicolumn{1}{|c|}{\multirow{-4}{*}{\textbf{1}}} & \multicolumn{1}{l|}{\textbf{MTFA + Both}} & \multicolumn{1}{r|}{{{4.31}}} & \multicolumn{1}{r|}{8.63} & \multicolumn{1}{r|}{\textcolor{black}{ \textbf{3.75}}} & \multicolumn{1}{r|}{{{22.15}}} & \multicolumn{1}{r|}{\textcolor{black}{ \textbf{6.25}}} & \multicolumn{1}{r|}{{{3.63}}} & \multicolumn{1}{r|}{\textcolor{black}{ \textbf{14.67}}} & \multicolumn{1}{r|}{\textcolor{black}{ \textbf{17.47}}} & \multicolumn{1}{r|}{{{22.16}}} & \multicolumn{1}{r|}{\textcolor{black}{ \textbf{11.62}}} & \textcolor{black}{ \textbf{18.15}} \\ \hline
\multicolumn{1}{|c|}{} & \multicolumn{1}{l|}{\textbf{Baseline MTFA}} & \multicolumn{1}{r|}{5.90} & \multicolumn{1}{r|}{8.87} & \multicolumn{1}{r|}{6.83} & \multicolumn{1}{r|}{33.04} & \multicolumn{1}{r|}{{{9.74}}} & \multicolumn{1}{r|}{3.10} & \multicolumn{1}{r|}{17.26} & \multicolumn{1}{r|}{19.25} & \multicolumn{1}{r|}{34.04} & \multicolumn{1}{r|}{\textcolor{black}{ \textbf{15.74}}} & 19.61 \\
\multicolumn{1}{|c|}{} & \multicolumn{1}{l|}{\textbf{MTFA + ITL}} & \multicolumn{1}{r|}{{{7.28}}} & \multicolumn{1}{r|}{{{11.22}}} & \multicolumn{1}{r|}{{{8.25}}} & \multicolumn{1}{r|}{32.31} & \multicolumn{1}{r|}{\textcolor{black}{ \textbf{10.72}}} & \multicolumn{1}{r|}{\textcolor{black}{ \textbf{6.83}}} & \multicolumn{1}{r|}{{{20.52}}} & \multicolumn{1}{r|}{{{22.69}}} & \multicolumn{1}{r|}{32.34} & \multicolumn{1}{r|}{14.88} & {{23.52}} \\
\multicolumn{1}{|c|}{} & \multicolumn{1}{l|}{\textbf{MTFA + IMS}} & \multicolumn{1}{r|}{6.95} & \multicolumn{1}{r|}{10.88} & \multicolumn{1}{r|}{7.75} & \multicolumn{1}{r|}{{{33.93}}} & \multicolumn{1}{r|}{7.49} & \multicolumn{1}{r|}{{{6.81}}} & \multicolumn{1}{r|}{19.84} & \multicolumn{1}{r|}{22.01} & \multicolumn{1}{r|}{\textcolor{black}{ \textbf{34.10}}} & \multicolumn{1}{r|}{15.04} & 21.47 \\
\multicolumn{1}{|c|}{\multirow{-4}{*}{\textbf{2}}} & \multicolumn{1}{l|}{\textbf{MTFA + Both}} & \multicolumn{1}{r|}{\textcolor{black}{ \textbf{7.60}}} & \multicolumn{1}{r|}{\textcolor{black}{ \textbf{11.58}}} & \multicolumn{1}{r|}{\textcolor{black}{ \textbf{8.78}}} & \multicolumn{1}{r|}{\textcolor{black}{ \textbf{34.13}}} & \multicolumn{1}{r|}{8.23} & \multicolumn{1}{r|}{6.70} & \multicolumn{1}{r|}{\textcolor{black}{ \textbf{23.87}}} & \multicolumn{1}{r|}{\textcolor{black}{ \textbf{25.92}}} & \multicolumn{1}{r|}{{{34.07}}} & \multicolumn{1}{r|}{{{15.13}}} & \textcolor{black}{ \textbf{27.18}} \\ \hline
\multicolumn{1}{|c|}{} & \multicolumn{1}{l|}{\textbf{Baseline MTFA}} & \multicolumn{1}{r|}{5.84} & \multicolumn{1}{r|}{8.98} & \multicolumn{1}{r|}{6.29} & \multicolumn{1}{r|}{34.56} & \multicolumn{1}{r|}{7.78} & \multicolumn{1}{r|}{4.31} & \multicolumn{1}{r|}{19.13} & \multicolumn{1}{r|}{21.83} & \multicolumn{1}{r|}{35.80} & \multicolumn{1}{r|}{15.93} & 21.09 \\
\multicolumn{1}{|c|}{} & \multicolumn{1}{l|}{\textbf{MTFA + ITL}} & \multicolumn{1}{r|}{7.49} & \multicolumn{1}{r|}{{{11.51}}} & \multicolumn{1}{r|}{8.23} & \multicolumn{1}{r|}{\textcolor{black}{ \textbf{38.45}}} & \multicolumn{1}{r|}{8.61} & \multicolumn{1}{r|}{{{6.38}}} & \multicolumn{1}{r|}{\textcolor{black}{ \textbf{24.88}}} & \multicolumn{1}{r|}{\textcolor{black}{ \textbf{27.52}}} & \multicolumn{1}{r|}{\textcolor{black}{ \textbf{38.55}}} & \multicolumn{1}{r|}{{{17.66}}} & {{27.44}} \\
\multicolumn{1}{|c|}{} & \multicolumn{1}{l|}{\textbf{MTFA + IMS}} & \multicolumn{1}{r|}{{{7.55}}} & \multicolumn{1}{r|}{11.45} & \multicolumn{1}{r|}{{{8.50}}} & \multicolumn{1}{r|}{{{38.07}}} & \multicolumn{1}{r|}{\textcolor{black}{ \textbf{9.21}}} & \multicolumn{1}{r|}{5.70} & \multicolumn{1}{r|}{{{24.20}}} & \multicolumn{1}{r|}{{{27.29}}} & \multicolumn{1}{r|}{{{38.50}}} & \multicolumn{1}{r|}{\textcolor{black}{ \textbf{18.10}}} & \textcolor{black}{ \textbf{27.56}} \\
\multicolumn{1}{|c|}{\multirow{-4}{*}{\textbf{3}}} & \multicolumn{1}{l|}{\textbf{MTFA + Both}} & \multicolumn{1}{r|}{\textcolor{black}{ \textbf{7.94}}} & \multicolumn{1}{r|}{\textcolor{black}{ \textbf{12.07}}} & \multicolumn{1}{r|}{\textcolor{black}{ \textbf{9.11}}} & \multicolumn{1}{r|}{37.97} & \multicolumn{1}{r|}{{{8.96}}} & \multicolumn{1}{r|}{\textcolor{black}{ \textbf{6.68}}} & \multicolumn{1}{r|}{23.77} & \multicolumn{1}{r|}{26.74} & \multicolumn{1}{r|}{38.32} & \multicolumn{1}{r|}{17.64} & 26.58 \\ \hline
\multicolumn{1}{|c|}{} & \multicolumn{1}{l|}{\textbf{Baseline MTFA}} & \multicolumn{1}{r|}{5.84} & \multicolumn{1}{r|}{9.13} & \multicolumn{1}{r|}{6.04} & \multicolumn{1}{r|}{35.44} & \multicolumn{1}{r|}{8.17} & \multicolumn{1}{r|}{4.22} & \multicolumn{1}{r|}{19.67} & \multicolumn{1}{r|}{22.96} & \multicolumn{1}{r|}{35.94} & \multicolumn{1}{r|}{\textcolor{black}{ \textbf{14.16}}} & 22.58 \\
\multicolumn{1}{|c|}{} & \multicolumn{1}{l|}{\textbf{MTFA + ITL}} & \multicolumn{1}{r|}{{{9.76}}} & \multicolumn{1}{r|}{{{14.37}}} & \multicolumn{1}{r|}{{{11.12}}} & \multicolumn{1}{r|}{\textcolor{black}{ \textbf{40.05}}} & \multicolumn{1}{r|}{\textcolor{black}{ \textbf{8.82}}} & \multicolumn{1}{r|}{{{9.89}}} & \multicolumn{1}{r|}{25.93} & \multicolumn{1}{r|}{29.28} & \multicolumn{1}{r|}{\textcolor{black}{ \textbf{40.05}}} & \multicolumn{1}{r|}{12.53} & {{30.32}} \\
\multicolumn{1}{|c|}{} & \multicolumn{1}{l|}{\textbf{MTFA + IMS}} & \multicolumn{1}{r|}{\textcolor{black}{ \textbf{10.36}}} & \multicolumn{1}{r|}{\textcolor{black}{ \textbf{16.27}}} & \multicolumn{1}{r|}{\textcolor{black}{ \textbf{11.79}}} & \multicolumn{1}{r|}{39.32} & \multicolumn{1}{r|}{8.08} & \multicolumn{1}{r|}{\textcolor{black}{ \textbf{11.36}}} & \multicolumn{1}{r|}{\textcolor{black}{ \textbf{26.34}}} & \multicolumn{1}{r|}{\textcolor{black}{ \textbf{30.30}}} & \multicolumn{1}{r|}{39.35} & \multicolumn{1}{r|}{12.37} & \textcolor{black}{ \textbf{30.91}} \\
\multicolumn{1}{|c|}{\multirow{-4}{*}{\textbf{5}}} & \multicolumn{1}{l|}{\textbf{MTFA + Both}} & \multicolumn{1}{r|}{9.39} & \multicolumn{1}{r|}{14.19} & \multicolumn{1}{r|}{10.42} & \multicolumn{1}{r|}{{{39.37}}} & \multicolumn{1}{r|}{{{8.36}}} & \multicolumn{1}{r|}{9.28} & \multicolumn{1}{r|}{{{26.16}}} & \multicolumn{1}{r|}{{{30.16}}} & \multicolumn{1}{r|}{{{39.50}}} & \multicolumn{1}{r|}{{{13.98}}} & 30.23 \\ \hline
\end{tabular}}
\end{table*}

\subsection{Results and Discussion}
\label{result_experiment}

{\color{black}
\textbf{State-of-the-art comparison.}
To prove the effectiveness of our proposed methods, we conducted experiments on our proposed CAMO-FS dataset. We tested with $K = \{1, 2, 3, 5\}$ shots, respectively. Since several recent work have not published their source code \cite{wang2022dynamic, han2023reference}, we adopted the typical models addressing both detection and instance segmentation tasks to compare with our proposed methods.
\textcolor{red}{Table} \ref{table:sota_comparison} presents the evaluation of the performance of our methods of instance triplet loss and memory storage over our baseline MTFA \cite{Ganea_2021_CVPR}, the model of Mask R-CNN \cite{Kaiming-ICCV2017} with sigmoid classifier, and the state-of-the-art method iFS-RCNN \cite{nguyen2022ifs} in the approach of few-shot instance segmentation. We reported experiments on those models and chose the common COCO-80 ResNet-101 as their base model to apply our proposed methods. The details of this decision are declared in the ablation section. 
In terms of instance segmentation, we improved over MTFA \cite{Ganea_2021_CVPR}, Mask RCNN$^\dagger$ \cite{Kaiming-ICCV2017}, and iFS-RCNN \cite{nguyen2022ifs}  by getting average AP values of $6.23 \%, 8.04 \%, 7.13\%$, respectively thanks to instance triplet loss, and $7.35 \%, 7.96 \%, 6.25 \%$, respectively thanks to instance memory storage. 
Regarding object detection, our FS-CDIS got average amounts of $7.14 \%, 7.34 \%, 6.96 \%$, respectively with instance triplet loss and $7.34 \%, 7.45 \%, 5.88 \%$, respectively with instance memory storage. 
The detailed performance of our methods is in \textcolor{red}{Table} \ref{table:sota_comparison}.
Despite the limited results, we defeated the very early models on detection and instance segmentation tasks on camouflaged images.

\textbf{Proposed modules evaluation.}
In \textcolor{red}{Table} \ref{table:improvement_result}, we also present the results of the baseline MTFA \cite{Ganea_2021_CVPR} with its original default configuration along with our proposed improvements.
On top of the baseline MTFA \cite{Ganea_2021_CVPR}, we establish fine-tuning configuration on this model by training all heads of classification, box regression, and mask prediction on few-shot novel data.
The reported results prove the performance of \hlc[white]{the proposed instance triplet loss, instance memory storage, and the combination of both loss functions.}
}

\begin{table}[!t]
\caption{Ablation study on the base model with 1-shot results. The best performances are marked in {\color{black}\textbf{bold}}. ``Triplet" stands for Instance Triplet Loss and ``Memory" stands for Instance Memory Storage.}
\label{table:ablation}
\centering
\adjustbox{width=\linewidth}{
\begin{tabular}{|l|l|c|c|c|c|c|c|}
\hline
\multirow{2}{*}{\textbf{Method}} & \multicolumn{1}{c|}{\multirow{2}{*}{\textbf{Base Model}}} & \multicolumn{3}{c|}{\textbf{Segmentation}} & \multicolumn{3}{c|}{\textbf{Detection}} \\ 
 & &  \multicolumn{1}{c|}{\textbf{AP}} & \multicolumn{1}{c|}{\textbf{AP50}} & \textbf{AP75} & \multicolumn{1}{c|}{\textbf{AP}} & \multicolumn{1}{c|}{\textbf{AP50}} & \textbf{AP75} \\ \hline
 \multirow{4}{*}{\textbf{Triplet}} & COCO-80 R-101 & {{\color{black}\textbf{4.46}}} & {{\color{black}\textbf{8.21}}} & {\color{black}\textbf{4.60}} & {{\color{black}\textbf{4.04}}} & {{\color{black}\textbf{8.65}}} & {\color{black}\textbf{2.98}} \\ 
 & COCO-80 R-50 & {3.68} & {{6.79}} & 3.81 & {2.85} & {{6.67}} & 1.65 \\ 
 & COCO-60 R-101 & {{3.87}} & {6.26} & {{3.90}} & {{3.37}} & 6.51 & {{2.69}} \\ 
 & COCO-60 R-50 & {2.56} & {4.25} & 2.79 & {2.28} & {4.13} & 2.26 \\ \hline
\multirow{4}{*}{\textbf{Memory}} & COCO-80 R-101 & {\color{black}\textbf{5.46}} & {\color{black}\textbf{9.20}} & {\color{black}\textbf{6.17}} & {\color{black}\textbf{4.50}} & {\color{black}\textbf{9.14}} & {\color{black}\textbf{3.45}} \\ 
 & COCO-80 R-50 & {{3.87}} & {{6.81}} & {{3.91}} & {{3.40}} & {{6.94}} & 2.76 \\ 
 & COCO-60 R-101 & {2.89} & {4.50} & 3.26 & {2.76} & {4.66} & {{2.81}} \\ 
 & COCO-60 R-50 & {2.63} & {4.50} & 3.02 & {2.25} & {4.50} & 1.65 \\ \hline
\end{tabular}}
\end{table}

In general, our approaches achieve outstanding results in comparison with the baseline.  Our improvements surpass MTFA by a remarkable margin. 
These results manifest the efficiency of our methods in the context of few-shot camouflaged instance segmentation. Both loss functions enhance the discrimination between foreground and background features which strongly supports the model to segment pixels that belong to the camouflaged animals. Regarding the memory storage and the triplet loss function, the results of the memory loss function are higher than those of the triplet loss function by about $1\%$. We realize that storing representatives for each class is a crucial element in few-shot learning. This technique not only expands the variants during training but also increases the consistency per class, so thereby model can segment difficult objects better. In these ways, we also improve the corresponding results in camouflage object detection.

In \textcolor{red}{Table} \ref{table:improvement_result}, our improvements help the model segment animals in various sizes. Specifically, all three metrics including APs, APm, and APl improve in comparison with the baseline model, which demonstrates that our model well segments small, medium, and large animals. This situation also happens in the detection task. When data is very scarce as in a 1-shot or 2-shot setting, the instance triplet loss function has comparative results with the instance memory storage function. However, in the context of 3-shot or 5-shot settings, the instance memory storage demonstrates outstanding efficiency thanks to storing and updating the memory via iterations to create discriminative features on a global level.
\hlc[white]{Actually, our proposed instance triplet loss is designed to differentiate the features among the foreground and background of a single camouflaged instance. Meanwhile, the instance memory storage aims to store features of multiple instances of the same category to enhance the general features. Thus, when combining the two approaches at the same time to train our FS-CDIS, the performance of the model fluctuates but still follows a trend. It can be seen from the reported numbers that with $K = \{1, 2, 3\}$ shots, the results of the combined loss seem to dominate the results of each separate component. However, notably around $K = 5$ shots, the AP, AP50, and AP75 of the instance memory storage yield better performance due to the information increase by more shots.}

In terms of quantitative comparison, \textcolor{red}{Figure} \ref{fig:visualization} illustrates the qualitative comparison among the results of 5-shot settings of the baseline MTFA \cite{Ganea_2021_CVPR} and our proposed methods of Instance Triplet Loss and Instance Memory Storage. 
We chose to visualize the images with a confidence threshold of $0.5$, which released a huge number of predictions with low confidence from the models. The four final rows indicate exemplary cases that either FS-CDIS-ITL or FS-CDIS-IMS can figure out camouflaged instances compared to the baseline. \hlc[white]{In these cases, our methods seem to be better the baseline although there are some imperfect cases where the prediction mask in the sixth row overlays irrelevant parts of the object (over-segmentation) or the results in the fourth, fifth, and last rows only capture some main parts of objects (under-segmentation). We conjecture the reasons for that is the image contains background clutter and the extremely vague boundary between foreground and background. To alleviate the phenomenon, it is feasible to further apply the post-processing methods on the output segmentation masks.}


\begin{table*}[!t]
\caption{{\color{black}Ablation study on the margin and the $\alpha$ ratio of the instance triplet loss in 1-shot settings. The best performances are marked in {\color{black}\textbf{bold}}.}}
\label{table:ablation_triplet}
\centering
\adjustbox{max width=0.99\textwidth}{
\begin{tabular}{|c|cclll|cclll|}
\hline
\textbf{AP} & \multicolumn{5}{c|}{\textbf{Segmentation}} & \multicolumn{5}{c|}{\textbf{Detection}} \\ \hline
\diagbox{$\textbf{\alpha}$}{\textbf{Margin}} & {\textbf{0}} & {\textbf{0.25}} & {\textbf{0.50}} & {\textbf{0.75}} & {\textbf{1.00}} & {\textbf{0}} & {\textbf{0.25}} & {\textbf{0.50}} & {\textbf{0.75}} & {\textbf{1.00}} \\ \hline 
1 & {3.89} & {4.50} & {3.92} & {\color{black}\textbf{5.16}} & 4.43 & {3.34} & {3.65} & {3.22} & {\color{black}\textbf{4.22}} & {{3.68}} \\ \hline
$1 \times 10^{-1}$ & {\color{black}\textbf{4.82}} & {{4.74}} & {4.46} & {{4.58}} & {\color{black}\textbf{4.57}} & {\color{black}\textbf{4.36}} & {\color{black}\textbf{4.27}} & {{4.04}} & {{4.16}} & {\color{black}\textbf{3.79}} \\ \hline
$1 \times 10^{-2}$ & {{4.29}} & {\color{black}\textbf{4.74}} & {\color{black}\textbf{4.69}} & {4.46} & {{4.39}} & {{4.02}} & {{3.97}} & {\color{black}\textbf{4.24}} & {4.06} & 3.71 \\ \hline
\end{tabular}}
\end{table*}

\textbf{Base model ablation study.}
We also conduct ablation experiments on different backbone base models of the COCO settings including general and few-shot concepts.
To be detailed, we report the performance of our proposed method of instance triplet loss and instance memory storage over four different backbones. The considered backbones are ResNet-50 and ResNet-101 \cite{he2016deep}. The two base datasets are MS-COCO with 80 classes and 60 classes, respectively. Thus, it led to the combination of four different base models (i.e. COCO-80/60 R-101, COCO-80/60 R-50).
As can be seen from \textcolor{red}{Table} \ref{table:ablation}, the performance of applying COCO-80 R-101 base weight yields better results among others evaluated on AP, AP@50, and AP@75 in both segmentation and detection tasks. In both cases of our two proposed improvements, the ablation results demonstrate our selection of COCO-80 R-101 is the best among the tested backbones of the base phase. For the segmentation task, we achieve $4.46$ and $5.46$ of AP reported for triplet loss and memory storage, respectively. For the detection task, we reach $4.04$ and $4.50$ also of AP, respectively. In summary, the chosen backbone of the base weight presents a higher performance of around $1\%$ to $2\%$ of evaluated on common metrics as in the table. To be explained, the base from COCO-80 contains more semantic concepts in comparison with the COCO-60 base, which leads to the higher performance reported. Note that all the results in this ablation section are reported for the 1-shot setting.

{\color{black}
\textbf{Ablation on instance triplet loss component.}
In terms of the instance triplet loss described in \textcolor{red}{Eq.} \ref{eq1}, we establish ablation experiments to evaluate the performance of the model with different configurations of margin and $\alpha$ value \hlc[white]{when the instance memory storage component is disabled (i.e. $\beta=0$)}. To this end, we set up the margin varying from $0$ to $1$, with a step of $0.25$. For the $\alpha$ ratio of the loss function (\textcolor{red}{Eq.} \ref{eq4}), we check out $\alpha$ = \{$1, 1 \times 10^{-1}, 1 \times 10^{-2}$\}. To be enhanced, the margin value indicates how distinguished foreground and background features are. Meanwhile, the $\alpha$ controls the effect of the instance triplet loss on the total loss of the framework. \textcolor{red}{Table} \ref{table:ablation_triplet} presents the evaluation of both detection and segmentation issues in 1-shot manner. As can be inferred from the table, the effect of $\alpha$ decides which margin should be selected for the triplet loss. With $\alpha = 1$ meaning we keep the original ratio of the loss, the segmentation result in 1-shot setting yields the highest performance of $5.16\%$ mAP with a $0.75$ margin value. Meanwhile, the detection result gets the highest performance of $4.36\%$ with $\alpha = 1 \times 10^{-1}$ and zero margin. This table offers a better understanding of the impact of $\alpha$ and the margin over the total performance.


\begin{table}[!t]
\caption{{\color{black}Ablation study on the capacity of the instance memory storage in 1-shot settings. The best performances are marked in {\color{black}\textbf{bold}}.}}
\label{table:ablation_memo_capacity}
\centering
\adjustbox{max width=0.5\textwidth}{
\begin{tabular}{|c|c|c|c|c|c|c|}
\hline
\multirow{2}{*}{\textbf{Capacity}} & \multicolumn{3}{c|}{\textbf{Segmentation}} & \multicolumn{3}{c|}{\textbf{Detection}} \\ 
 & \multicolumn{1}{c|}{\textbf{AP}} & \multicolumn{1}{c|}{\textbf{AP50}} & \textbf{AP75} & \multicolumn{1}{c|}{\textbf{AP}} & \multicolumn{1}{c|}{\textbf{AP50}} & \textbf{AP75} \\ \hline
32 & {4.56} & {7.30} & 5.02 & {3.85} & {7.72} & 2.91 \\ \hline
64 & {4.51} & {{7.67}} & 4.49 & {3.94} & {{8.37}} & 2.84 \\ \hline
128 & {4.53} & {7.55} & 4.87 & {4.13} & {7.98} & 3.62 \\ \hline
256 & {4.56} & {7.50} & 5.02 & {4.01} & {8.22} & 3.39 \\ \hline
512 & {{\color{black}\textbf{4.76}}} & {7.57} & {\color{black}\textbf{5.37}} & {{\color{black}\textbf{4.48}}} & {8.25} & {\color{black}\textbf{4.44}} \\ \hline
1024 & {{4.72}} & {{\color{black}\textbf{8.06}}} & {{5.20}} & {{4.14}} & {{\color{black}\textbf{8.45}}} & {{3.79}} \\ \hline
\end{tabular}}
\end{table}

\textbf{Ablation on instance memory storage component.}
As for the instance memory storage, as introduced in \textcolor{red}{Eq.} \ref{eq:infonce}, and \textcolor{red}{Eq.} \ref{eq4}, there are several parameters that need analyzing, listed as the amount of capacity in the memory storage and the $\beta$ ratio controlling the effect of the memory storage loss in the total loss. \textcolor{red}{Table} \ref{table:ablation_memo_capacity}, and \textcolor{red}{Table} \ref{table:ablation_memo_beta} present the ablation experimental results of those issues \hlc[white]{when the instance triplet loss function is disabled (i.e. $\alpha=0$)}, respectively. 
In terms of the capacity of the memory storage, we establish experiments on a range of memory capacity of $2^i$ where $i=\{5, 6, 7, 8, 9, 10\}$. The reported results figure out that the performance on both segmentation and detection tasks increases with a larger capacity of memory storage. To be detailed, with a capacity of $512$, the mAP metric achieves the highest value among configurations, i.e. $4.76\%$ and $4.48\%$ for segmentation and detection, respectively. Empirically, we select $512$ to be the suitable capacity of the memory storage, not the largest. To this end, the larger capacity can confuse the model in the process of learning when retrieving information in such a large memory bank. 
Besides, \textcolor{red}{Table} \ref{table:ablation_memo_beta} expresses the effectiveness of the memory loss to the total loss function. As can be inferred, $\beta = 1 
\times 10^{-4}$ gives the best performance evaluated on mAP, AP50, and AP75 among all configurations. 


\begin{table}[!t]
\caption{Ablation study on the $\beta$ ratio of the instance memory loss ({\color{red}Eq.} \ref{eq4}) in 1-shot settings. The best performances are marked in {\color{black}\textbf{bold}}.}
\label{table:ablation_memo_beta}
\centering
\adjustbox{max width=0.55\textwidth}{
\begin{tabular}{|l|c|c|c|c|c|c|}
\hline
\multicolumn{1}{|c|}{\multirow{2}{*}{$\textbf{\beta}$}} & \multicolumn{3}{c|}{\textbf{Segmentation}} & \multicolumn{3}{c|}{\textbf{Detection}} \\ 
\multicolumn{1}{|c|}{} & \multicolumn{1}{c|}{\textbf{AP}} & \multicolumn{1}{c|}{\textbf{AP50}} & \textbf{AP75} & \multicolumn{1}{c|}{\textbf{AP}} & \multicolumn{1}{c|}{\textbf{AP50}} & \textbf{AP75} \\ \hline
$1 \times 10^{-1}$ & {3.36} & {6.58} & 2.91 & {3.69} & {8.02} & 2.90 \\ \hline
$1 \times 10^{-2}$ & {{4.57}} & {{8.02}} & {{4.74}} & {3.73} & {7.78} & 2.76 \\ \hline
$1 \times 10^{-3}$ & {4.51} & {7.15} & 4.67 & {3.87} & {7.16} & 3.51 \\ \hline
$1 \times 10^{-4}$ & {{\color{black}\textbf{5.12}}} & {{\color{black}\textbf{8.71}}} & {\color{black}\textbf{5.54}} & {{\color{black}\textbf{4.58}}} & {{\color{black}\textbf{9.23}}} & {\color{black}\textbf{3.69}} \\ \hline
$1 \times 10^{-5}$ & {4.44} & {7.58} & 3.89 & {{4.06}} & {{7.99}} & {{3.63}} \\ \hline
\end{tabular}}
\end{table}

\section{Conclusion}
\label{conclusion}
In this work, we investigated the interesting yet challenging problem of few-shot learning for camouflaged animal detection and segmentation. We first collect a new dataset, CAMO-FS, for benchmarking purposes. We then propose a novel method to efficiently detect and segment the camouflaged animals in the images. In particular, we introduce the instance triplet loss and the instance memory storage. The extensive experiments demonstrated that our proposed method achieves state-of-the-art performance on the newly constructed dataset. We expect our work will encourage more research work in this field. In the future, we would like to extend our work with more shots for new classes. In addition, we aim to improve the computational model by taking the context into consideration.

\begin{figure*}[!h]
    \centering
    \includegraphics[width=0.92\textwidth]{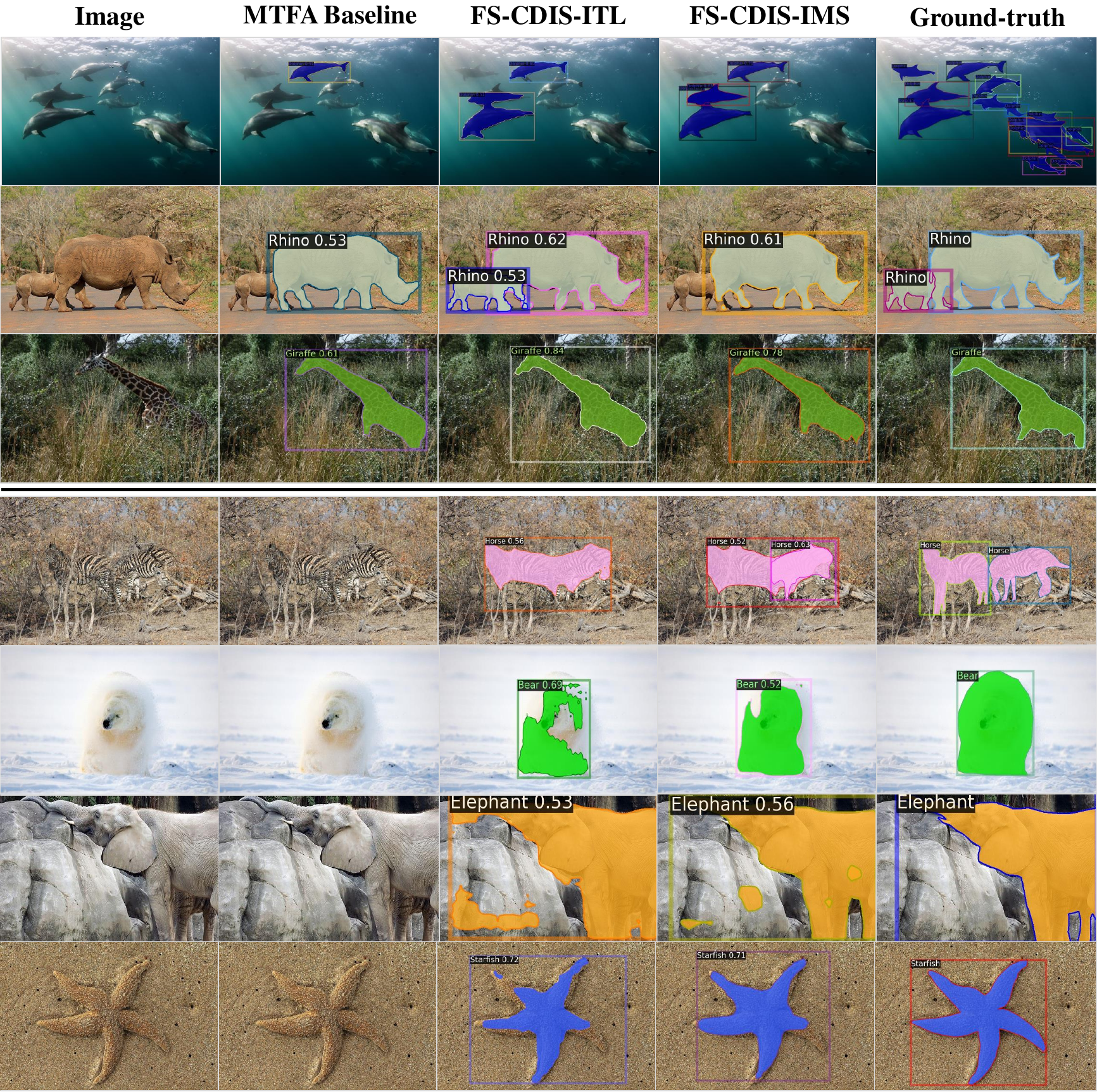}
    \caption{{\color{black}Qualitative comparison among the selected baseline MTFA \cite{Ganea_2021_CVPR} and our proposed methods. The results are from 5-shot settings. Predicted images are visualized with a confidence threshold of $0.5$, which released a huge number of predictions with low confidence from the models. The four final rows indicate exemplary cases that either FS-CDIS-ITL or FS-CDIS-IMS can figure out camouflaged instances compared to the baseline.}}
    \label{fig:visualization}
\end{figure*}

\section*{Acknowledgment}
This research was supported by The VNUHCM-University of Information Technology's Scientific Research Support Fund.

\bibliographystyle{IEEEtran}
\bibliography{bibtex}

\begin{IEEEbiography}
    [{\includegraphics[width=1in,height=1.25in,clip,keepaspectratio]
    {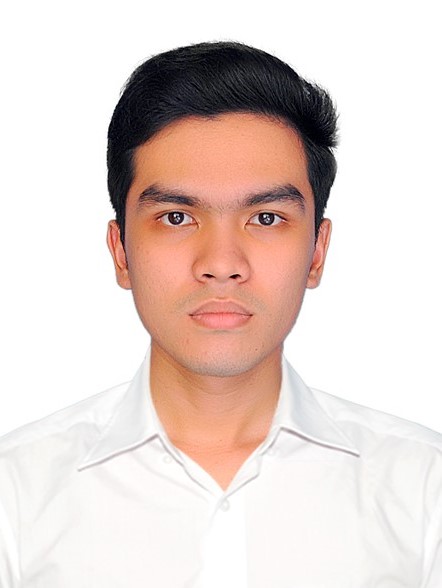}}]{Thanh-Danh Nguyen} is currently a Ph.D. student at the Laboratory of Multimedia Communications, University of Information Technology, VNU-HCM. He received his M.Sc. and B.Sc. degree (Hons.) in computer science from University of Information Technology, VNU-HCM, in 2024 and 2021, respectively. His research interests are computer vision and deep learning.
\end{IEEEbiography}

\begin{IEEEbiography}
    [{\includegraphics[width=1in,height=1.25in,clip,keepaspectratio]
    {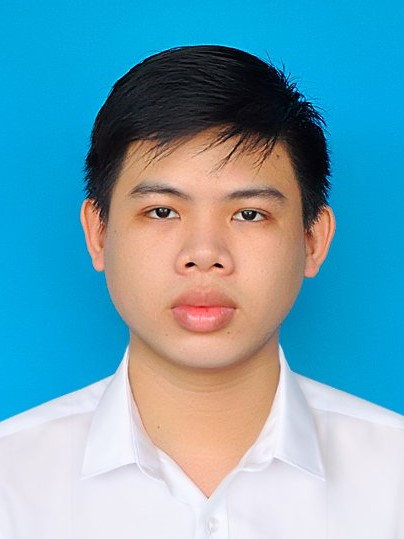}}]{Anh-Khoa Nguyen Vu} received M.Sc. and B.Sc. degrees in computer science from University of Information Technology, VNU-HCM in 2023 and 2021, respectively. His current research interests include few-shot learning, computer vision, and deep learning.
\end{IEEEbiography}

\begin{IEEEbiography}
    [{\includegraphics[width=1in,height=1.25in,clip,keepaspectratio]
    {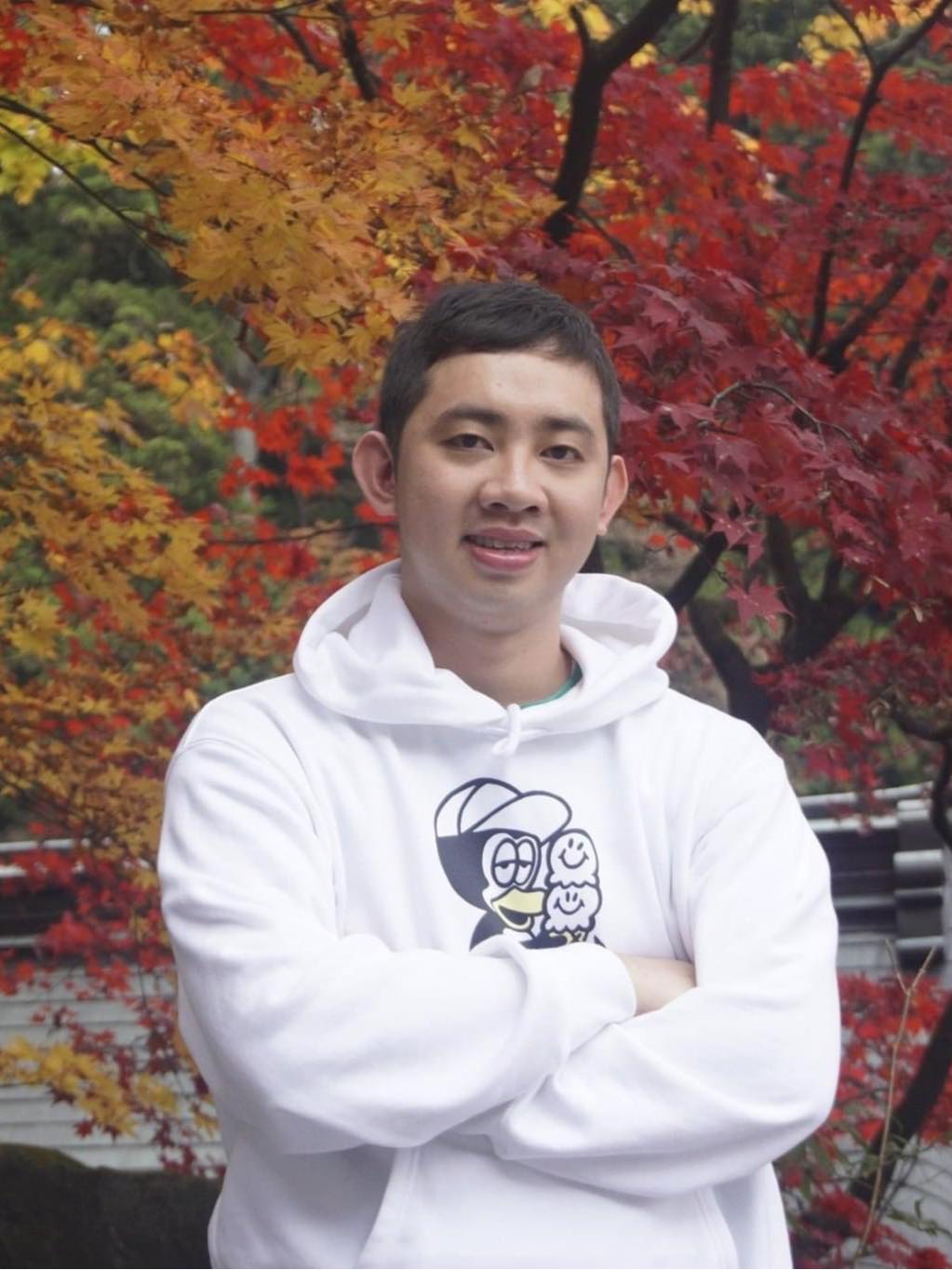}}]{Nhat-Duy Nguyen} is a former researcher at the Laboratory of Multimedia Communications, University of Information Technology, VNU-HCM. He obtained his M.Sc. degree and B.Sc degree (Hons.) from University of Information Technology, VNU-HCM. His research interests are computer vision and machine learning, especially in object detection and segmentation. He also worked as a lecturer at the faculty of computer science. He is currently a software engineer at the FPT corporation.

\end{IEEEbiography}

\begin{IEEEbiography}
    [{\includegraphics[width=1in,height=1.25in,clip,keepaspectratio]
    {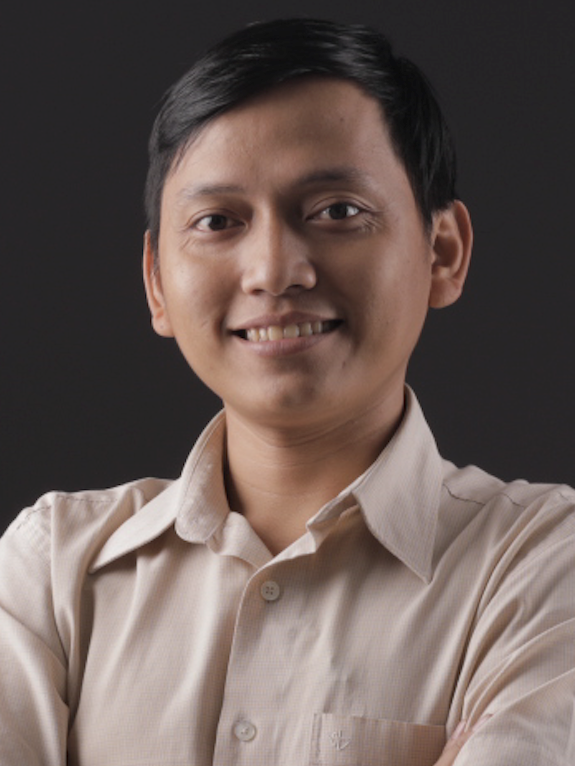}}]{Vinh-Tiep Nguyen} is currently a lecturer at University of Information Technology, Head of the Laboratory of Multimedia Communications (MMLab), VNU-HCM. He obtained his Ph.D. degree from University of Information Technology, VNU-HCM, and M.Sc. degree from University of Science, VNU-HCM in a co-program with John von Neumann Institute, VNU-HCM. His research interests are computer vision and machine learning. He has a great passion for transferring knowledge and research skills to his students.
\end{IEEEbiography}

\begin{IEEEbiography}
    [{\includegraphics[width=1in,height=1in,clip,keepaspectratio]
    {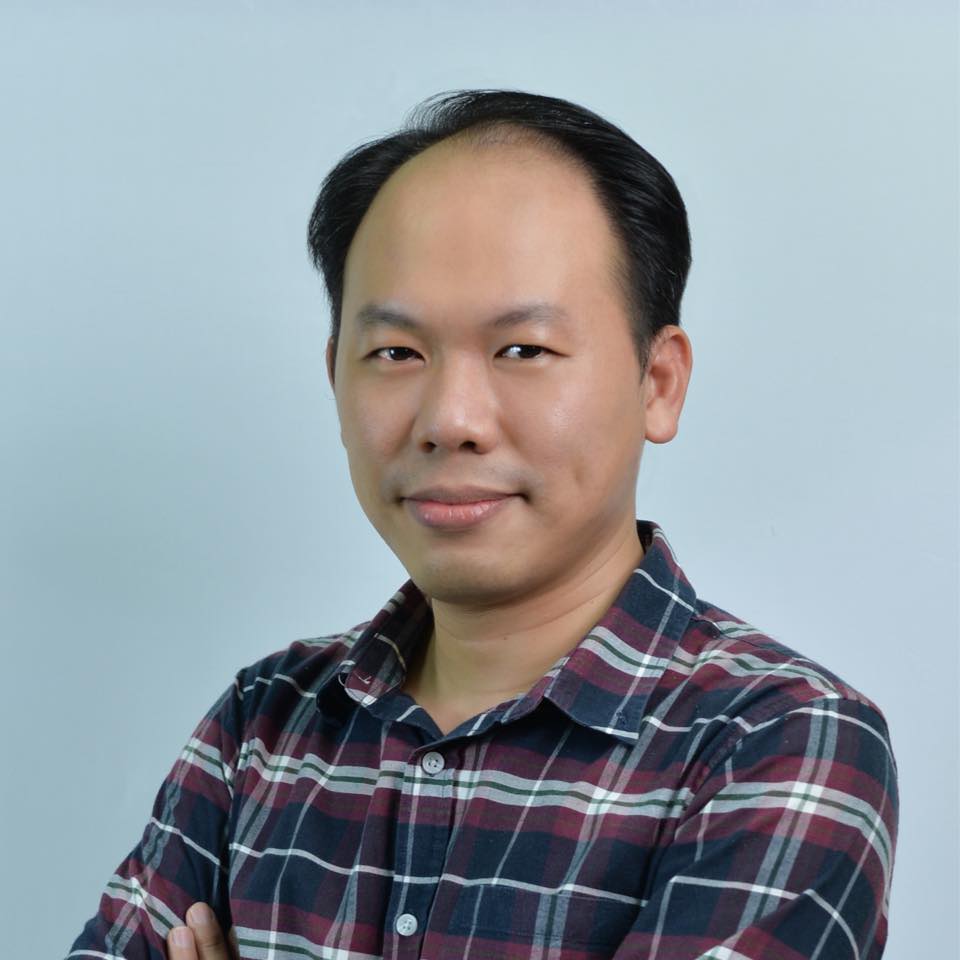}}]{Thanh Duc Ngo} is a lecturer in the Faculty of Computer Science, University of Information Technology, VNU-HCM where he has been since 2014. He completed his Ph.D. at The Graduate University for Advanced Studies (SOKENDAI) in 2013. His research interests lie in the areas of Computer Vision and Multimedia Content Analysis.

\end{IEEEbiography}

\begin{IEEEbiography}
    [{\includegraphics[width=1in,height=1.25in,clip,keepaspectratio]
    {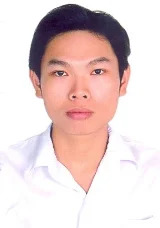}}]{Thanh-Toan Do} is a Senior Lecturer at the Faculty of Information Technology, Monash University. He obtained his Ph.D. in computer science at the French National Institute for Research in Computer Science and Control (INRIA) in 2012. From 2013 to 2016, he was a Research Fellow at the Singapore University of Technology and Design. From 2016 to 2018, he was a Research Fellow at the Australian Centre for Robotic Vision and the University of Adelaide. From 2018 to 2020, he was a Lecturer at the University of Liverpool. His research interests include computer vision and machine learning. 

\end{IEEEbiography}

\begin{IEEEbiography}
    [{\includegraphics[width=1in,height=1.25in,clip,keepaspectratio]
    {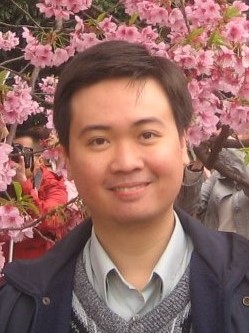}}]{Minh-Triet Tran} obtained his B.Sc., M.Sc., and Ph.D. degrees in computer science from University of Science, VNU-HCM, in 2001, 2005, and 2009. He joined the University of Science, VNU-HCM, in 2001. His research interests include cryptography and security, computer vision and machine learning, and human-computer interaction. He was a visiting scholar at National Institutes of Informatics (NII, Japan) in 2008, 2009, and 2010, and at University of Illinois at Urbana-Champaign (UIUC) in 2015-2016. He is currently the Vice Rector of University of Science, VNU-HCM. He is also the Head of Software Engineering Laboratory and the Deputy Head of Artificial Intelligence Laboratory, University of Science, VNU-HCM. He is a member of the Management Board of Vietnam Information Security Association (South Branch) and also a member of the Executive Committee of ICT Program for Smart Cities (2018-2020) of Ho Chi Minh City.
\end{IEEEbiography}

\begin{IEEEbiography}
    [{\includegraphics[width=1in,height=1.25in,clip,keepaspectratio]
    {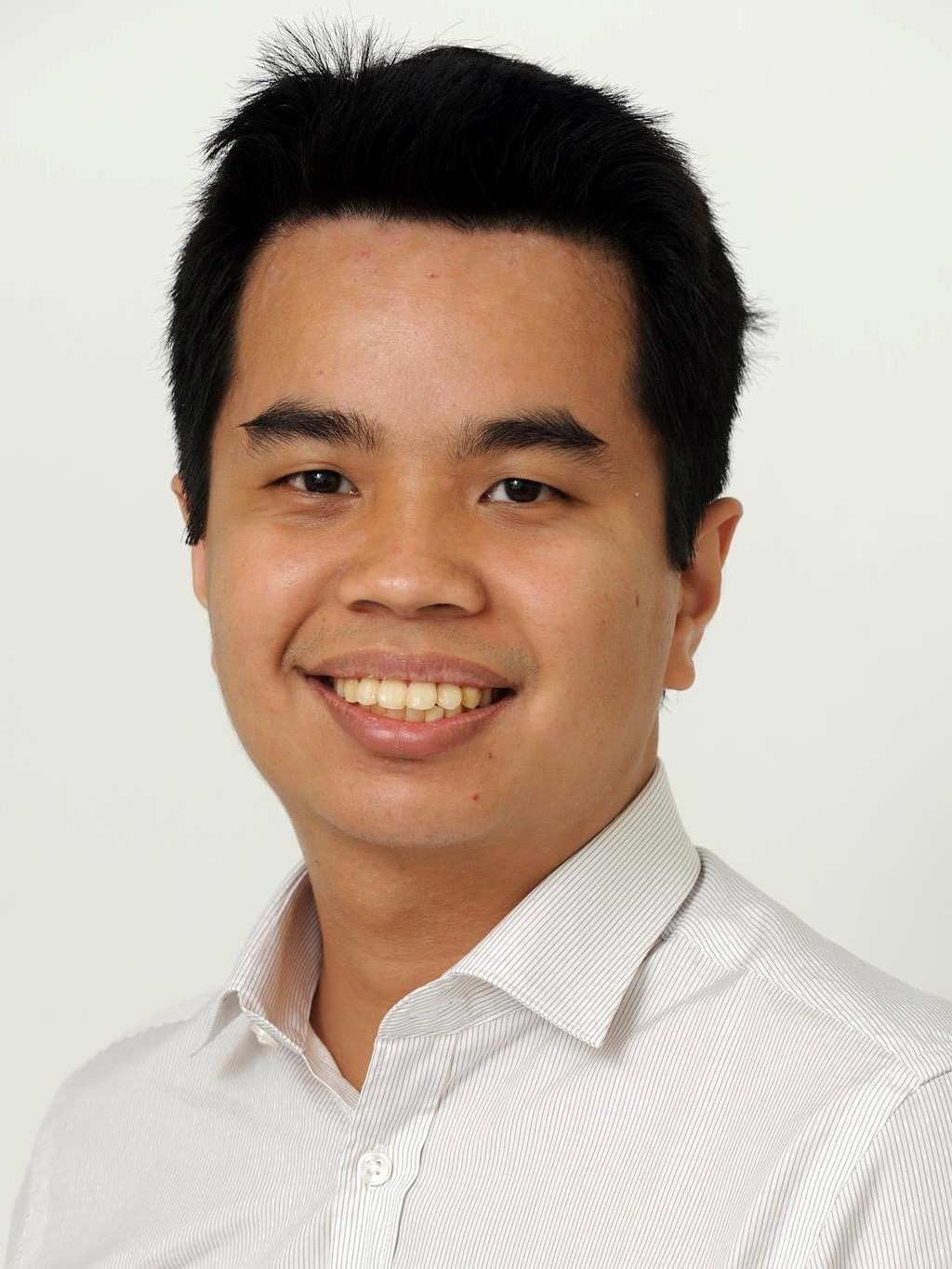}}]{Tam V. Nguyen}  is an Associate Professor at Department of Computer Science, University of Dayton. He received his Ph.D. degree from the National University of Singapore in 2013. His research topics include artificial intelligence, computer vision, deep learning, multimedia content analysis, and mixed reality.He is one of the pioneers of camouflage analysis research. He has authored and co-authored more than 120 papers with more than 3,000 citations. His H-index is 29. He is an IEEE Senior Member.
\end{IEEEbiography}

\balance
\EOD

\end{document}